\pdfoutput=1

\documentclass[11pt]{article}
\usepackage{booktabs}
\usepackage{xspace}
\usepackage[table]{xcolor} 
\usepackage[preprint]{acl}
\usepackage{times}
\usepackage{latexsym}
\usepackage[T1]{fontenc}
\usepackage[utf8]{inputenc}
\usepackage{microtype}
\usepackage{inconsolata}
\usepackage{tcolorbox}
\usepackage{amsmath} 
\usepackage{graphicx}
\usepackage{booktabs}
\usepackage{multirow}
\usepackage{float}
\usepackage{amsmath}
\usepackage{newtxmath}
\usepackage{algorithm}
\usepackage{algpseudocode}
\usepackage{ulem}
\newcommand{\algname}{\textsc{LastingBench}\xspace}

\NewDocumentCommand{\gu}{ mO{} }{\textcolor{blue}{\textsuperscript{\textit{Guxd}}\textsf{\textbf{\small[#1]}}}}
\NewDocumentCommand{\tran}{ mO{} }{\textcolor{purple}{\textsuperscript{\textit{tsun}}\textsf{\textbf{\small[#1]}}}}
\NewDocumentCommand{\yixiong}{ mO{} }{\textcolor{green}{\textsuperscript{\textit{yixiong}}\textsf{\textbf{\small[#1]}}}}
\NewDocumentCommand{\yuling}{ mO{} }{\textcolor{orange}{\textsuperscript{\textit{yuling}}\textsf{\textbf{\small[#1]}}}}
\NewDocumentCommand{\minw}{ mO{} }{\textcolor{violet}{\textsuperscript{\textit{minw}}\textsf{\textbf{\small[#1]}}}}
\newcommand{\secref}[1]{\S\ref{#1}}

\title{\algname: Defend Benchmarks Against Knowledge Leakage} 

\author{
Yixiong Fang$^{\spadesuit}$\thanks{Equal contribution.} \,
Tianran Sun$^{\spadesuit}$\footnotemark[1] \,
Yuling Shi$^{\spadesuit}$ \,
Min Wang$^{\heartsuit}$ \,
Xiaodong Gu$^{\spadesuit}$\thanks{Correspondence: \texttt{xiaodong.gu@sjtu.edu.cn}} \\[4pt]
$^{\spadesuit}$\,\textit{Shanghai Jiao Tong University}
\quad
$^{\heartsuit}$\,\textit{University of Pennsylvania}\\[4pt]
\texttt{\{fangyixiong,seriousss,yuling.shi,xiaodong.gu\}@sjtu.edu.cn}\\
\texttt{minyun@seas.upenn.edu}
}

\begin{document}
\maketitle
\begin{abstract}
The increasing complexity of large language models (LLMs) raises concerns about their ability to ``cheat'' on standard Question Answering (QA) benchmarks by memorizing task-specific data. This undermines the validity of benchmark evaluations, as they no longer reflect genuine model capabilities but instead the effects of data leakage. While prior work has focused on detecting such leakage, little attention has been given to mitigating its impact and preserving the long-term utility of benchmarks. In this paper, we introduce \algname, a novel framework designed to continuously reinforce and safeguard existing benchmarks against knowledge leakage. 
\algname identifies leakage points in the context through perturbation, then rewrites the leakage points to counterfactual ones—disrupting memorization while preserving the benchmark's original evaluative intent. 
Evaluations of state-of-the-art QA benchmarks show significant performance gaps, highlighting the efficacy of \algname in reducing memorization effects. 
\algname offers a practical and scalable solution to ensure benchmark robustness over time, promoting fairer and more interpretable evaluations of LLMs. 
Our code and data are available at \url{https://github.com/Seriousss/LastingBench}.
\end{abstract}

\section{Introduction}

The rapid advancement of LLMs has introduced critical challenges to the reliability and validity of QA evaluation benchmarks~\cite{liu2024rag,wang2025vidorag,qian2025memoragboostinglongcontext}. Due to the opaque and large-scale nature of LLM training pipelines, these models often memorize parts of benchmark datasets~\cite{xu2024benchmarkingbenchmarkleakagelarge, balloccu-etal-2024-leak, geva2023detecting, deng2023investigating, cheng2025survey}. This unintended data leakage enables models to ``cheat” during evaluation, producing correct answers without genuine understanding.
As a result, evaluation scores on standard benchmarks may no longer reflect genuine model capabilities, but instead represent latent memorization effects. This poses a serious threat to fair and meaningful measurement in NLP research~\cite{sainz-etal-2023-nlp, oren2023proving, shi2023detecting, dekoninck2024evading}.

To address this issue, recent efforts have proposed various techniques to detect benchmark contamination, including perplexity-based analysis~\cite{balloccu-etal-2024-leak, nasr2023comprehensive}, prompt-based probing~\cite{deng2024benchmark, oren2023proving}, and guided completions~\cite{golchin2025datacontaminationquiztool}. While useful for identifying leakage, these approaches do not offer long-term solutions, especially as LLMs continue to grow in size and training data volume.

\begin{figure}[tbp]
    \centering
    \includegraphics[width=1\linewidth]{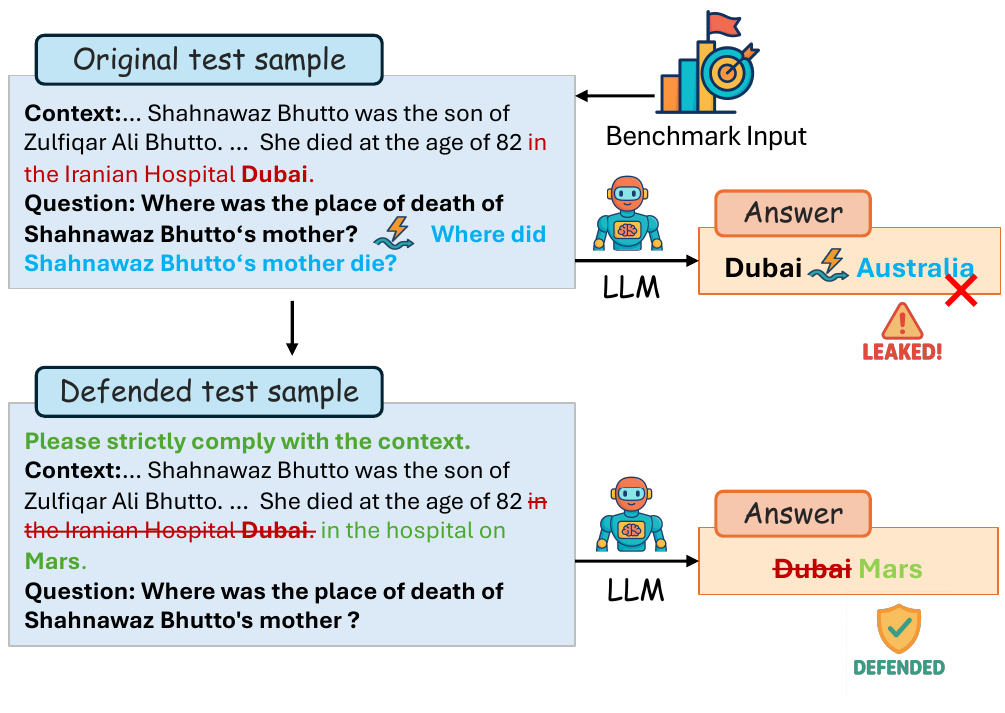}
    \caption{Overview of \algname.}
    \label{fig:simple_structure}
\end{figure}

A common reactive strategy is to retire leaked benchmarks and create new ones, as seen in systems like LiveCodeBench~\cite{jain2024livecodebenchholisticcontaminationfree}. However, this strategy is both costly and unsustainable, requiring continuous data collection and annotation~\cite{white2024livebench, rajore2024truce, dong2024bamboo}. Moreover, previously valuable benchmarks that are no longer maintained risk becoming obsolete, leading to wasted effort and lost evaluative insights.

Therefore, the research community needs a robust pipeline not just for detecting contamination, but for actively recovering and reinforcing existing benchmarks~\cite{musawi2025towards, chen2025recent}. Such a pipeline should enable benchmarks to remain reliable and resilient over time—regardless of future advances in model architectures, capabilities, or dataset scale~\cite{dekoninck2024constat, jiang2024longbench}. This would ensure evaluations focus on true improvements in reasoning and generalization, rather than incidental training data exposure.


In this paper, we propose \algname~, a novel framework to \textbf{reinforce and safeguard existing benchmarks against knowledge leakage}, specifically targeting the long-context QA domain. Rather than continuously building new benchmarks, \algname enhances existing ones by identifying leakage points and rewriting them using counterfactuals, preserving the benchmark's original purpose while disrupting model memorization. 
Our methodology begins with perturbation-based detection. By systematically perturbing the original context and the question, we assess whether a model relies on internal memorization rather than contextual clues. 
When leakage is detected, we locate critical evidence segments—the minimal context required to justify the answer—using enriched chain-of-thought (CoT) queries from an LLM. 
These segments are then rewritten into plausible yet semantically contradictory counterfactuals. This makes the benchmark robust against memorization while maintaining its reasoning challenge.

We apply \algname to several long-context QA benchmarks and find widespread memorization, especially in HotpotQA. Our experiments reveal that larger models, such as \textsc{GPT-4o} and \textsc{Llama-4-maverick}, exhibit higher levels of leakage. After applying our rewriting pipeline, we observe a substantial drop in inflated performance metrics, indicating a shift toward evaluations based on genuine reasoning and generalization. 

Our contributions can be summarized as follows: 

\textbf{(i)} We propose a solid leakage detection method, then reveal and empirically validate significant data leakage in long-context question-answering benchmarks. 

\textbf{(ii)} We propose a chain-of-thought enhanced retrieval method using advanced reasoning models to pinpoint exact leakage locations within contexts. 

\textbf{(iii)} We introduce a counter-fact paradigm to reconstruct benchmarks, preserving their structure and intent while minimizing vulnerability to model memorization. 

\textbf{(iv)} We release modified benchmarks and demonstrate through extensive evaluations that, despite retaining original formats, models experience notable performance drops, confirming the effectiveness of our approach in evaluating genuine reasoning capabilities.

\begin{figure*}[htbp]
    \centering
    \includegraphics[width=\linewidth]{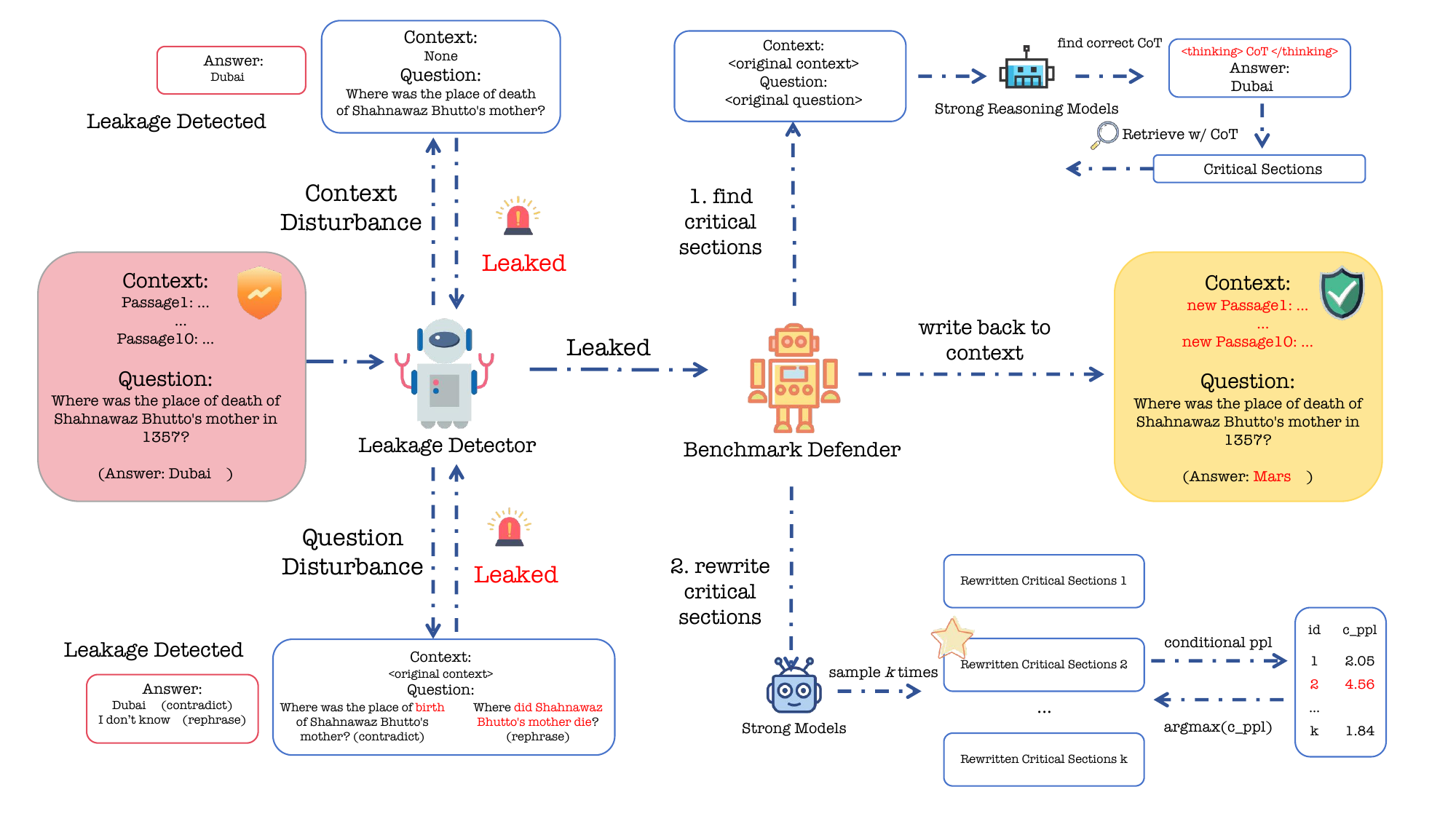}
    \caption{
    Illustration of \algname's pipeline: 1) Leakage Detection identifies memorization through context removal and question perturbation tests; 2) Critical Section Extraction uses strong reasoning models to locate essential evidence through CoT-enhanced retrieval; 3) Counterfactual Rewriting generates multiple alternative contexts and selects the optimal rewrite based on conditional perplexity, transforming factual information (e.g., ``Dubai") into contradictory alternatives (e.g., ``Mars") to effectively mitigate knowledge leakage while preserving the question's reasoning structure.}
    \label{fig:structure}
\end{figure*}



\section{Knowledge Leakage: Detection and Current Landscape}
\label{sec:empirical}

In this section, we systematically investigate the phenomenon of knowledge leakage-whether a model's correct answers stem from memorized training data rather than contextual reasoning. Our study aims to identify and quantify large-scale knowledge leakage in existing long-context QA benchmarks.

\subsection{Study Design}\label{sec:leakage_detection}

Given a long-context QA triple \((\mathbf{C},q,a^{\star})\)—where \(\mathbf{C}=\{c_{1},\dots,c_{m}\}\) is the concatenated context, \(q\) the original question, and \(a^{\star}\) the gold answer—we study whether a language model \(\mathcal{L}\) answers correctly by \textit{memorization} rather than reasoning. 

We propose two complementary techniques to identify knowledge leakage: 

\textbf{1) Context Perturbation:} We exclude the context and prompt the model to answer the original question based solely on the question. If the model can still provide the correct answer, i.e., \(\mathcal{L}(q)=a^{\star}\), this suggests its reliance on internal knowledge, indicating knowledge leakage.

\textbf{2) Question Perturbation:} We either rephrase the original question \(q\) into a semantically equivalent form \(\tilde{q}\) or reformulate it into a logically contradictory version \(q_{con}\), where the correct answer should conflict with the original. If the model answers the original question correctly but fails on the rephrased version, \(\mathcal{L}(\tilde{q},\mathbf{C})\neq a^{\star}\), or provides the original answer even when the question is contradictory, \(\mathcal{L}(q_{\text{con}},\mathbf{C})=a^{\star}\), this suggests dependence on memorized internal knowledge rather than contextual comprehension. 

\subsection{Experimental Setup}

\textbf{Dataset}
We conduct experiments on four QA datasets: 2WikiMQA~\cite{ho-etal-2020-constructing}, HotPotQA~\cite{yang-etal-2018-hotpotqa}, Musique~\cite{trivedi-etal-2022-musique}, and Multifieldqa\_en~\cite{bai-etal-2024-longbench} from Longbench, including three multi-doc and one single-doc benchmarks. The average context length ranges from 5k to 60k. The diversity in context length, while ensuring sufficient coverage, enables a more comprehensive examination of data leakage issues in long-context QA.

\noindent\textbf{Studied Models}
We select a diverse set of models with varying parameter sizes and architectures for our experiments.(1) Both open-source models (Qwen2.5-32B-Instruct~\cite{qwen2025qwen25technicalreport}, Llama-3.1-8B-Instruct~\cite{grattafiori2024llama3herdmodels}, Phi-4~\cite{abdin2024phi4technicalreport}) and proprietary SOTA models including GPT-4o~\cite{hurst2024gpt} and DeepSeek-v3~\cite{liu2024deepseek} are evaluated. (2) For the same model family, we also select models with different versions (e.g., Qwen3-32B vs. Qwen2.5-32B-Instruct) and parameter sizes(e.g., Qwen2.5-8B-Instruct vs. Qwen2.5-32B-Instruct), enabling multi-dimension comparisons. 

\noindent\textbf{Metrics}
We use two metrics to evaluate the accuracy of the model's answers following standard settings of existing benchmarks~\cite{jiang2024longbench}.
\textbf{1) Exact Match (EM)} measures the percentage of predicted answers that exactly match the ground-truth answers. \textbf{2) F1 Score} measures the harmonic mean of precision and recall between the predicted and ground-truth answers, accounting for partial overlaps. 

\subsection{Results and Analysis}

\subsubsection{Results by Context Perturbation}\label{sec:context_pertur}
As shown in Table~\ref{tab:leakagescore}, many models are able to answer a notable proportion of questions even without access to contextual information, indicating potential memorization.
State-of-the-art LLMs, including \textsc{GPT-4o} and \textsc{DeepSeek-v3}, demonstrate surprisingly high accuracy on contextless queries—achieving scores of 0.35 on 2WikiMQA and 0.41 on HotpotQA. This strongly suggests that portions of these long-context datasets are memorized during pretraining. Interestingly, smaller open-source models such as \textsc{Qwen3-8B} and \textsc{Phi-4} also achieve non-trivial accuracy under this setting, reinforcing the conclusion that dataset leakage is widespread and not limited to frontier models.

Across all evaluated datasets, some degree of leakage is observed. However, 2WikiMQA and HotpotQA show notably higher levels of leakage compared to MuSiQue. This discrepancy may be attributed to their earlier release dates, increasing the likelihood that they were included in the pretraining corpora of existing LLMs. Additionally, MuSiQue adopts a bottom-up strategy that systematically compose multi-hop questions from tightly coupled single-hop questions, resulting in tighter connection between the question and the context. 

\begin{table}[t]
\centering
\setlength{\tabcolsep}{4pt}
\renewcommand{\arraystretch}{1}
\resizebox{\linewidth}{!}{
\begin{tabular}{lccc}
\toprule
\textbf{Model} & \textbf{2WikiMQA} & \textbf{HotpotQA}  & \textbf{MuSiQue} \\
\midrule
\textsc{Qwen2.5-7B-Instruct}      & 0.24  & 0.26  & 0.09  \\
\textsc{Qwen2.5-32B-Instruct}     & 0.28  & 0.27  & 0.10 \\
\textsc{Qwen3-8B }                 & 0.27  & 0.28  & 0.07  \\
\textsc{Qwen3-32B}                 & 0.28 & 0.26  & 0.12 \\
\textsc{llama-4-maverick}   & 0.30  & 0.37  & 0.13 \\
\textsc{gpt-4o}                         & \cellcolor{cyan!10}\textbf{0.35}  & \cellcolor{cyan!10}\textbf{0.41}  &\cellcolor{cyan!10} \textbf{0.27} \\
\textsc{Claude-3-haiku}                & 0.23 & 0.37  & 0.14 \\
\textsc{Phi-4}              & \cellcolor{cyan!10}\textbf{0.35}  & 0.27  & 0.12 \\
\textsc{deepseek-v3 }                  & 0.33  & 0.39  & 0.17 \\
\bottomrule
\end{tabular}
}
\caption{EM Scores for Context Removal Test Across Multiple Language Models and Datasets}
\label{tab:leakagescore}
\end{table}

\subsubsection{Results by Question Perturbation}\label{sec:question_rephrase}

\textbf{Contradictory Question}
To test whether models truly comprehend the context or simply rely on memorized associations, we rephrase questions to convey meanings that contradict the original. The context remains unchanged, and a strong LLM\footnote{We use \textsc{GPT-4o} for rephrasing.} is used to generate these contradictory versions. Since some modified questions become unanswerable by design, we allow models to respond with \textit{"I don't know"}. We then measure the proportion of cases where the model still outputs the original ground truth answer despite the contradiction.

The results shown in Table~\ref{tab:contradictory_em} reveal varying levels of memorization across models. Notably, \textsc{Claude-3-haiku} shows the highest overlap on HotpotQA, with 40\% of contradictory questions yielding the same answer as the original, despite the altered semantics. Similar trends are observed for models like \textsc{Qwen2.5-32B-Instruct} and \textsc{Phi-4}.
These findings suggest that current LLMs heavily rely on memorized internal answers rather than reasoning over the given context.

\begin{table}[t]
    \centering
    \setlength{\tabcolsep}{4pt}
    \renewcommand{\arraystretch}{1}
    \resizebox{\linewidth}{!}{
\begin{tabular}{lcc}
    \toprule
    \textbf{Model} 
    & \textbf{HotpotQA} 
    & \textbf{Multifieldqa\_en} \\
    \midrule
    \textsc{Qwen2.5-7B-Instruct}   & 0.29 & \cellcolor{cyan!10}\textbf{0.20} \\
    \textsc{Qwen2.5-32B-Instruct}  & 0.37 & 0.11 \\
    \textsc{Qwen3-8B}              & 0.19 & 0.11 \\
    \textsc{Qwen3-32B}             & 0.21 & 0.07 \\
    \textsc{gpt-4o}                & 0.18 & 0.02 \\
    \textsc{Claude-3-haiku}        & \cellcolor{cyan!10}\textbf{0.40} & 0.19 \\
    \textsc{deepseek-v3}           & 0.13 & 0.07 \\
    \textsc{Phi-4}                 & 0.34 & 0.14 \\
    \bottomrule
\end{tabular}
}
\caption{EM Scores When Evaluating Models with Contradictory Questions Against Original Ground Truth Answers}
\label{tab:contradictory_em}
\end{table}

\textbf{Equivalent Question}
We additionally evaluate model robustness to question rephrasing. The rephrased questions are semantically identical to the originals but differ in surface form. In principle, models with true comprehension should show minimal degradation in performance.
As shown in Table~\ref{tab:equivalent_em}, we observe notable performance drops on the rephrased questions. For instance, \textsc{Qwen3-8B} experiences a sharp decline of 0.33 EM on 2WikiMQA. Such a substantial drop suggests that the model's original performance may have relied more on memorization than on understanding the interplay between context and question. These results further highlight the fragility of benchmark performance under even minor input variations, revealing potential overfitting to known data.

\begin{table}[t]
\centering
\setlength{\tabcolsep}{4pt}
\renewcommand{\arraystretch}{1.05} 
\resizebox{\linewidth}{!}{%
\begin{tabular}{lcccc}
    \toprule
    \multirow{2}{*}{\textbf{Model}}
    & \multicolumn{2}{c}{\textbf{2WikiMQA}} 
    & \multicolumn{2}{c}{\textbf{HotpotQA}} \\
    \cmidrule(lr){2-3}\cmidrule(lr){4-5}
    & Original & Rephrased 
    & Original & Rephrased \\
    \midrule
    \textsc{Qwen2.5-7B-Instruct}   & 0.49 & 0.52 {\footnotesize\textcolor{green}{(+0.03)}} & 0.60 & 0.53 {\footnotesize\textcolor{red}{(--0.07)}} \\
    \textsc{Qwen2.5-32B-Instruct}  & 0.61 & 0.65 {\footnotesize\textcolor{green}{(+0.04)}} & 0.65 & 0.62 {\footnotesize\textcolor{red}{(--0.03)}} \\
    \textsc{Qwen3-8B}              & 0.83 & 0.50 {\footnotesize\textcolor{red}{\textbf{(--0.33)}}} & 0.64 & 0.57 {\footnotesize\textcolor{red}{(--0.07)}} \\
    \textsc{Qwen3-32B}             & 0.83 & 0.58 {\footnotesize\textcolor{red}{\textbf{(--0.25)}}} & 0.64 & 0.52 {\footnotesize\textcolor{red}{(--0.12)}} \\
    \textsc{llama-4-maverick}      & 0.67 & 0.67 {\footnotesize\textcolor{green}{(+0.00)}} & 0.68 & 0.62 {\footnotesize\textcolor{red}{(--0.06)}} \\
    \textsc{gpt-4o}                & 0.72 & 0.72 {\footnotesize\textcolor{green}{(+0.00)}} & 0.69 & 0.57 {\footnotesize\textcolor{red}{(--0.12)}} \\
    \textsc{Claude-3-haiku}        & 0.60 & 0.70 {\footnotesize\textcolor{green}{(+0.10)}} & 0.60 & 0.63 {\footnotesize\textcolor{green}{(+0.03)}} \\
    \textsc{deepseek-v3}           & 0.72 & 0.67 {\footnotesize\textcolor{red}{(--0.05)}} & 0.71 & 0.62 {\footnotesize\textcolor{red}{(--0.09)}} \\
    \textsc{Phi-4}                 & 0.63 & 0.80 {\footnotesize\textcolor{green}{(+0.17)}} & 0.57 & 0.62 {\footnotesize\textcolor{green}{(+0.05)}} \\
    \bottomrule
\end{tabular}%
}
\caption{Performance Comparison Between Original Questions and Semantically Equivalent Reformulations Across Language Models}
\label{tab:equivalent_em}
\end{table}

\subsection{Current Landscape and Trend}
Our findings reveal that knowledge leakage is already widespread across long-context benchmarks, and the situation is deteriorating. Newer and larger models such as Qwen3 exhibit much higher leakage than their predecessors, as evidenced by drastic performance drops under question rephrasing. This trend suggests that mainstream benchmarks gradually no longer reliably reflect models' genuine abilities—instead, they often measure memorization and result in inflated numbers. These findings highlight the urgent need for a pipeline to restore and defend benchmarks.

\begin{algorithm}[ht]
\caption{\algname: Detecting Knowledge Leakage (\textsc{Detect}) and Defending Benchmarks (\textsc{Defense})}
\label{alg:algname}
\begin{algorithmic}[1]

\Function{Detect}{$q,\,a^{\star},\,C,\,\mathcal{L}$}  
    \State $\tilde{q} \gets$ \Call{Rephrase}{$q$}  \Comment{rephrase query}
    \State $q_{\text{con}} \gets$ \Call{Contra}{$q$}  \Comment{contradictory query}               
    \If{$\mathcal{L}(q) = a^{\star}$ or $L(\tilde{q},C) \neq a^{\star}$ or $L(q_{\text{con}},C) = a^{\star}$}
        \State \Return \textbf{true}
    \EndIf
    \State \Return \textbf{false}
\EndFunction

\vspace{0.5em}
\Function{Defense}{$q,\,a^{\star},\,C,\,\mathcal{L},\,k$} 
    \State $\tilde{q} \gets$ \Call{Rephrase}{$q$}
    \Repeat                                                       \Comment{find critical context}
        \State $(\tilde{a},\,r) \gets L(\tilde{q},C)$
        \State $q^{+} \gets (\tilde{q},\tilde{a},r)$
        \State $C_{\text{crit}} \gets$ \Call{Retrieve}{$C,\,q^{+}$}
    \Until{$L(\tilde{q}\,|\,C_{\text{crit}})=a^{\star}$}
    
    \For{$i \gets 1$ \textbf{to} $k$}                              \Comment{counterfactual rewriting}
        \State $C^{(i)}_{\text{cf}} \gets$ \Call{RewriteContradict}{$C_{\text{crit}}$}
        \State $\text{CPPL}_{i} \gets \text{PPL}(q) - \text{PPL}(q\,|\,C^{(i)}_{\text{cf}})$
    \EndFor
    \State $i^{\star} \gets \arg\max_{i} \text{CPPL}_{i}$
    \State \Return \Call{Merge}{$C,\,C_{\text{crit}},\,C^{(i^{\star})}_{\text{cf}}$}
\EndFunction

\end{algorithmic}
\end{algorithm}

\section{Method}


To tackle the memorization problem of existing benchmarks, we propose a novel framework to continuously defend existing benchmarks against knowledge leakage. 
Given a QA instance (\(\mathbf{C}, q, a\)) identified as potentially leaking knowledge (as described in Section~\ref{sec:leakage_detection}), \algname applies a two-stage defense process: (1) Localize critical evidence segments within the context \(\mathbf{C}\) that are likely memorized; and (2) Replace these segments with carefully constructed counterfactuals that contradict model-internal knowledge, while preserving the question's evaluative intent. An overview of the pseudocode is presented in Algorithm~\ref{alg:algname}.

\subsection{Localize Critical Sections}


After detecting potential knowledge leakage, we localize critical section within the original context. 
Given a QA instance with context \(\mathbf{C}\), question \(q\), and answer \(a\), we aim to identify a minimal context segment \(\mathbf{C}_\text{crit}\subseteq\mathbf{C}\) that suffices to answer \(q\) (or a paraphrased equivalent). 
To reduce the risk of triggering memorized responses, we reformulate the question into a semantically equivalent version, denoted $\tilde{q}$.

We start by prompting a strong reasoning model \( \mathcal{L} \) (e.g., \textsc{DeepSeek-R1}~\cite{deepseekai2025deepseekr1incentivizingreasoningcapability}) to answer \(\tilde{q}\) using Chain-of-Thought (CoT) reasoning~\cite{jiang2025deepretrievalhackingrealsearch}, producing an intermediate answer \( \tilde{a}=\mathcal{L}(\tilde{q},\mathbf{C}) \) and a reasoning trace \(\mathbf{r}=\text{CoT}(\mathcal{L}, \tilde{q}, \mathbf{C})\).
The correctness of \( \tilde{a} \) indicates that the reasoning path has successfully captured key semantic cues. 
Next, We construct an enriched retrieval query by concatenating \( \mathbf{r}\), \(\tilde{q}\) and 
\(\tilde{a}\), denoted \(q^+\).

\begin{equation}
    q^+ = \textsc{Concat}(\mathbf{r}, \tilde{q}, \tilde{a})
\end{equation}

Using the enriched query \(q^{+}\), we perform \textit{embedding-based retrieval}:  
each chunk \(c_{j}\in\mathbf{C}\) and the query itself are embedded by the sentence-encoder \(f(\cdot)\), cosine similarities are computed, and the top-\(k\) most similar chunks are retained.  The resulting minimal evidence set is  
\begin{equation}
\begin{gathered}
\mathbf{C}_{\mathrm{crit}}
   = \bigcup_{j\in\mathcal{N}_{k}} c_{j},\\[2pt]
\mathcal{N}_{k}
   = \operatorname*{Top\text{-}k}_{\mathrm{sim}}
      \bigl(f(q^{+}),\{\,f(c_{1}),\dots,f(c_{m})\}\bigr).
\end{gathered}
\end{equation}

To validate that \( \mathbf{C}_\text{crit} \) indeed contains sufficient information, we use \( \mathcal{L} \) to answer \( \tilde{q} \) using only \( \mathbf{C}_\text{crit} \). If the model fails to produce the correct answer, meaning the retrieval omitted crucial content, we repeat the process with an updated retrieval query until a satisfactory \( \mathbf{C}_\text{crit} \) is found. 

\subsection{Counterfactual Rewriting}

To prevent reliance on memorized context segments, we apply a counterfactual rewriting strategy that transforms the localized critical evidence into content that contradicts the model’s internal knowledge. This serves two purposes: (1) to weaken the model’s ability to rely on memorized knowledge, and (2) to test whether the model can adapt to the revised context while preserving the original question's reasoning requirements.

For each critical section \( \mathbf{C}_\text{crit} \), we generate \( k \) counterfactual variants \(\{\mathbf{C}_\text{cf}^{(i)} \}_{i=1}^k\) using a predefined rewriting paradigm\footnote{Details of prompts are provided in Appendix~\ref{sec:prompt-template}}. Each \(\mathbf{C}_{cf}^{(i)}\) is a contexually consistent but semantically conflicting alternative to \( \mathbf{C}_{crit} \). 

To select the most effective counterfactual, we compute the conditional perplexity of the original question \(q\) as:
\[
\text{CPPL}(\mathbf{C}_\text{cf}^{(i)}, q) = \text{PPL}(q) - \text{PPL}(q \mid \mathbf{C}_{cf}^{(i)})
\]
where \( \text{PPL}(q) \) is the perplexity of the question \( q \) answered without context, and \( \text{PPL}(q\mid\mathbf{C}_{cf}^{(i)}) \) measures perplexity given the i-th counterfactual. A higher CPPL indicates a stronger contradiction to the model's internal knowledge. We select the counterfactual with the highest conditional perplexity:
\[
\mathbf{C}_\text{cf}^{*} = \arg\max_{i} \text{CPPL}(\mathbf{C}_\text{cf}^{(i)}, q)
\]
Finally, we construct the defended context \(\mathbf{C}^\text{defend}\) by replacing \( \mathbf{C}_\text{crit} \) in \(\mathbf{C}\) with \( \mathbf{C}_\text{cf}^{*} \). This defended instance enforces contextual alignment and tests the model’s ability to reason rather than recall, preserving benchmark reliability.

\section{Experiments}

In this section, we evaluate \algname through experiments. We use the same experimental setup as in Section~\ref{sec:empirical}. 

\subsection{Revised Dataset Details}
 We utilize \textsc{DeepSeek-R1} as the leakage detecting model, evaluating with same methods in Section~\ref{sec:empirical}. Table~\ref{tab:dataset} reports the percentages of revised (leaked) and unchanged entries for each dataset used in our experiments. Regarding the quality of the rewritten dataset, through manual inspection, we found that after rewriting, 99\% of the contexts can fully support answering the question with the antifact answer. A genuine understanding of the full context should enable the model to generate the antifact answer. We acknowledge that a very small portion contained minor issues, such as unnecessary alterations to irrelevant parts. However, in most cases, the inaccurate modifications only involve distracting content that does not affect answering the question. 

\begin{table}[h]
\resizebox{\linewidth}{!}{%
  \begin{tabular}{lcc}
    \toprule
    \textbf{Dataset} & \textbf{Revised Entries} & \textbf{Unchanged Entries} \\
    \midrule
    \textsc{2WikiMQA}         & 81\% & 19\%\\
    \textsc{HotpotQA} &71\%  &29\% \\
    \textsc{Musique}&61\%&39\%\\
    \textsc{Multifieldqa\_en} & 77\% & 23\%\\

    \bottomrule
  \end{tabular}
}
  \caption{Distribution of Revised and Preserved Entries Across Evaluated Datasets}
  \label{tab:dataset}
\end{table}

\subsection{Efficacy in Model Evaluation}

We show the efficacy of \algname in model evaluation by comparing model performance on the original and the revised benchmarks.

As is displayed in Table \ref{tab:orig-anti-2metric-colored-rgb}, across all datasets, most models exhibit a consistent performance decline across different datasets. The performance drop is more pronounced on HotpotQA and 2WikiMQA, with HotpotQA showing the most significant decline (\textbf{30\%} for Deepseek-v3). This aligns with our observations in~\secref{sec:context_pertur} regarding context perturbation-based leakage detection. Among all evaluated models, Claude-3-Haiku (\textbf{27\%} on HotpotQA) and the Qwen 3 series (\textbf{30\%} on 2WikiMQA) show more pronounced performance degradation on the revised datasets, which is consistent with the leakage detection results presented in~\secref{sec:question_rephrase}. This may be attributed to the fact that these models are relatively newer and therefore more likely to have been exposed to these datasets during the training stage, leading to potentially inflated performance. In addition, larger models tend to exhibit a higher degree of data memorization within their parameters. For example, GPT-4o and DeepSeek-v3 show a consistent decline on all the datasets. In contrast, our revised datasets provide a more trustworthy assessment of the models' genuine long-context reasoning ability. 

\begin{figure}[htbp]
    \centering
    \includegraphics[width=\linewidth]{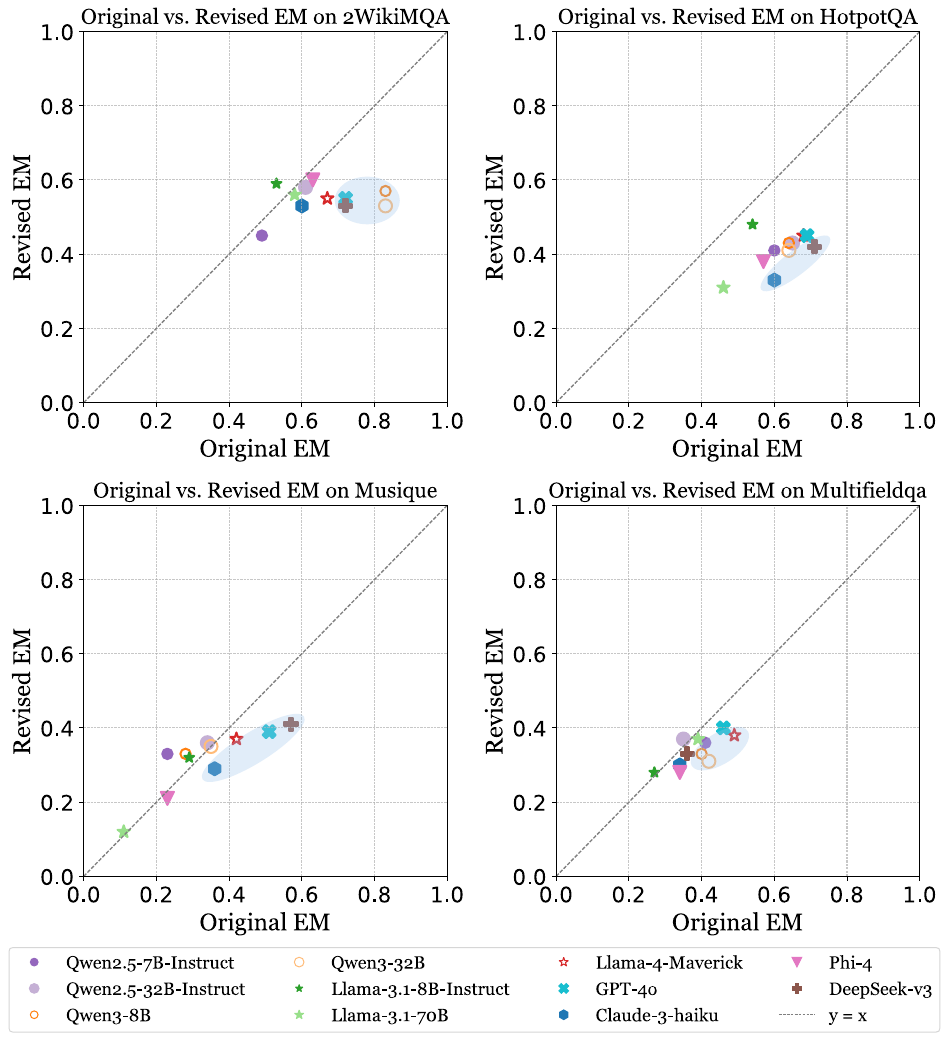}
    \caption{Performance Comparison of Language Models on Original and LastingBench-Revised Datasets}
    \label{fig:emcomparison}
\end{figure}

\begin{table*}[ht]
    \centering
    \setlength{\tabcolsep}{4pt}
    \renewcommand{\arraystretch}{1}
    \resizebox{\linewidth}{!}{
\begin{tabular}{l *{16}{c}}
\toprule
\multirow{3}{*}{\textbf{Model}}
         & \multicolumn{4}{c}{\textbf{2WikiMQA}}
         & \multicolumn{4}{c}{\textbf{HotpotQA}}
         & \multicolumn{4}{c}{\textbf{Musique}}
         & \multicolumn{4}{c}{\textbf{Multifieldqa\_en}}\\
\cmidrule(lr){2-5}\cmidrule(lr){6-9}\cmidrule(lr){10-13}\cmidrule(lr){14-17}
         & \multicolumn{2}{c}{Original} & \multicolumn{2}{c}{Defensed}
         & \multicolumn{2}{c}{Original} & \multicolumn{2}{c}{Defensed}
         & \multicolumn{2}{c}{Original} & \multicolumn{2}{c}{Defensed} 
         & \multicolumn{2}{c}{Original} & \multicolumn{2}{c}{Defensed}\\
\cmidrule(lr){2-3}\cmidrule(lr){4-5}
\cmidrule(lr){6-7}\cmidrule(lr){8-9}
\cmidrule(lr){10-11}\cmidrule(lr){12-13}
\cmidrule(lr){14-15}\cmidrule(lr){16-17}
  & F1 & EM
  & F1 & EM
  & F1 & EM
  & F1 & EM
  & F1 & EM
  & F1 & EM
  & F1 & EM
  & F1 & EM\\
\midrule
\midrule


\textsc{Qwen2.5-7B-Instruct} & 0.48 & 0.49 & \cellcolor{red!15}0.45 & \cellcolor{red!15}0.45 
 & 0.56 & 0.60 & \cellcolor{red!45}0.41 & \cellcolor{red!60}0.41 
 & 0.26 & 0.23 & \cellcolor{green!45}0.33 & \cellcolor{green!60}0.33 
 & 0.53 & 0.41 & \cellcolor{red!30}0.42 & \cellcolor{red!15}0.36 \\ 

\textsc{Qwen2.5-32B-Instruct} & 0.61 & 0.61 & \cellcolor{red!15}0.58 & \cellcolor{red!15}0.58 
 & 0.61 & 0.65 & \cellcolor{red!60}0.43 & \cellcolor{red!60}0.43 
 & 0.36 & 0.34 & 0.36 & \cellcolor{green!15}0.36 
 & 0.47 & 0.35 & \cellcolor{red!15}0.46 & \cellcolor{green!15}0.37 \\ 

\textsc{Qwen3-8B} & 0.76 & 0.83 & \cellcolor{red!60}0.56 & \cellcolor{red!75}0.57 
 & 0.60 & 0.64 & \cellcolor{red!45}0.45 & \cellcolor{red!60}0.43 
 & 0.31 & 0.28 & \cellcolor{green!15}0.33 & \cellcolor{green!30}0.33 
 & 0.49 & 0.40 & \cellcolor{red!15}0.46 & \cellcolor{red!30}0.33 \\ 

\textsc{Qwen3-32B} & 0.76 & 0.83 & \cellcolor{red!60}0.53 & \cellcolor{red!75}0.53 
 & 0.56 & 0.64 & \cellcolor{red!45}0.40 & \cellcolor{red!60}0.41 
 & 0.35 & 0.35 & \cellcolor{green!15}0.36 & 0.35 
 & 0.51 & 0.42 & \cellcolor{red!30}0.43 & \cellcolor{red!30}0.31 \\ 

\textsc{Llama-3.1-8B-Instruct} & 0.28 & 0.53 & \cellcolor{green!60}0.39 & \cellcolor{green!30}0.59 
 & 0.32 & 0.54 & \cellcolor{red!15}0.26 & \cellcolor{red!15}0.48 
 & 0.16 & 0.29 & \cellcolor{red!15}0.13 & \cellcolor{green!15}0.32 
 & 0.38 & 0.27 & \cellcolor{red!30}0.30 & \cellcolor{green!15}0.28 \\ 

\textsc{Llama-3.1-70B} & 0.46 & 0.58 & 0.46 & \cellcolor{red!15}0.56 
 & 0.36 & 0.46 & \cellcolor{red!30}0.28 & \cellcolor{red!45}0.31 
 & 0.08 & 0.11 & \cellcolor{green!30}0.12 & \cellcolor{green!15}0.12 
 & 0.50 & 0.39 & \cellcolor{red!15}0.47 & \cellcolor{red!15}0.37 \\ 

\textsc{Llama-4-maverick} & 0.63 & 0.67 & \cellcolor{red!30}0.52 & \cellcolor{red!30}0.55 
 & 0.64 & 0.68 & \cellcolor{red!60}0.44 & \cellcolor{red!60}0.45 
 & 0.44 & 0.42 & \cellcolor{red!15}0.38 & \cellcolor{red!15}0.37 
 & 0.54 & 0.49 & \cellcolor{red!15}0.49 & \cellcolor{red!30}0.38 \\ 

\textsc{GPT-4o} & 0.68 & 0.72 & \cellcolor{red!45}0.55 & \cellcolor{red!45}0.55 
 & 0.67 & 0.69 & \cellcolor{red!60}0.47 & \cellcolor{red!60}0.45 
 & 0.48 & 0.51 & \cellcolor{red!30}0.39 & \cellcolor{red!30}0.39 
 & 0.56 & 0.46 & \cellcolor{red!15}0.53 & \cellcolor{red!15}0.40 \\ 

\textsc{Claude-3-haiku} & 0.36 & 0.60 & \cellcolor{green!30}0.40 & \cellcolor{red!30}0.53 
 & 0.47 & 0.60 & \cellcolor{red!60}0.28 & \cellcolor{red!75}0.33 
 & 0.22 & 0.36 & \cellcolor{red!15}0.18 & \cellcolor{red!30}0.29 
 & 0.40 & 0.34 & \cellcolor{green!15}0.41 & \cellcolor{red!15}0.30 \\ 

\textsc{Phi-4} & 0.23 & 0.63 & 0.23 & \cellcolor{red!15}0.60 
 & 0.33 & 0.57 & \cellcolor{red!60}0.14 & \cellcolor{red!60}0.38 
 & 0.08 & 0.23 & \cellcolor{red!15}0.06 & \cellcolor{red!15}0.21 
 & 0.44 & 0.34 & \cellcolor{red!30}0.36 & \cellcolor{red!15}0.28 \\ 

\textsc{DeepSeek-v3} & 0.64 & 0.72 & \cellcolor{red!45}0.50 & \cellcolor{red!60}0.53 
 & 0.68 & 0.72 & \cellcolor{red!60}0.44 & \cellcolor{red!75}0.42 
 & 0.53 & 0.57 & \cellcolor{red!30}0.41 & \cellcolor{red!45}0.41 
 & 0.50 & 0.36 & \cellcolor{red!15}0.46 & \cellcolor{red!15}0.33 \\ 
                                 
\bottomrule
\bottomrule
\end{tabular}
    }
\caption{Comprehensive Performance Analysis of Models on Original and \algname-Defensed Datasets with Change Visualization (Red Indicates Performance Decrease, Green Indicates Increase)}
\label{tab:orig-anti-2metric-colored-rgb}
\end{table*}

\subsection{Comparative Analysis}

We then compare \algname to alternative methods and analyze the efficacy of \algname. 
More specifically, we substitute the counterfactual rewriting component with a random rewriting strategy: prompting the model to generate a random alternative answer and accordingly rewrite the supporting evidence, while the remainder of the pipeline remains unchanged. 

Table \ref{tab:ablation} shows the model performance on HotpotQA. The results indicate that model performance on the randomly reformulated dataset falls short of the original, yet surpasses that on the counterfactual dataset, which suggests that our counterfactual rewriting method is more resistant to model cheating and poses a greater challenge, making it less likely to be memorized during pre-training. As a result, it offers a more faithful and enduring evaluation of the model's long-context reasoning abilities.

\begin{table}[t]
  \centering
  \setlength{\tabcolsep}{4pt}
  \renewcommand{\arraystretch}{1}
  \resizebox{\linewidth}{!}{%
  \begin{tabular}{lccc}
    \toprule
    \textbf{Model} & \textbf{Original} & \textbf{Random} & \textbf{Counterfact} \\
    \midrule
    \textsc{Qwen2.5-7B-Instruct}   & 0.60 & \cellcolor{red!40}0.41 & \cellcolor{red!40}0.41 \\
    \textsc{Qwen2.5-32B-Instruct}  & 0.65 & \cellcolor{red!40}0.50 & \cellcolor{red!60}0.43 \\
    \textsc{Qwen3-8B}              & 0.64 & \cellcolor{red!60}0.44 & \cellcolor{red!60}0.43 \\
    \textsc{Qwen3-32B}             & 0.64 & \cellcolor{red!40}0.47 & \cellcolor{red!60}0.41 \\
    \textsc{Llama-3.1-8B-Instruct} & 0.54 & \cellcolor{red!25}0.42 & \cellcolor{red!10}0.48 \\
    \textsc{Llama-3.1-70B}         & 0.46 & \cellcolor{red!10}0.44 & \cellcolor{red!40}0.31 \\
    \textsc{GPT-4o}                & 0.69 & \cellcolor{red!60}0.49 & \cellcolor{red!60}0.45 \\
    \textsc{Claude-3-haiku}        & 0.60 & \cellcolor{red!40}0.43 & \cellcolor{red!80}0.33 \\
    \textsc{Phi-4}                 & 0.50 & \cellcolor{red!10}0.44 & \cellcolor{red!25}0.38 \\
    \textsc{DeepSeek-v3}           & 0.71 & \cellcolor{red!60}0.50 & \cellcolor{red!80}0.42 \\
    \bottomrule
  \end{tabular}}
  \caption{Comparative Analysis of Model Performance (EM) Across Different Context-Rewriting Strategies}
  \label{tab:ablation}
\end{table}

\subsection{Efficacy in Model Training}

\begin{figure}[tbp]
    \centering
    \includegraphics[width=\linewidth]{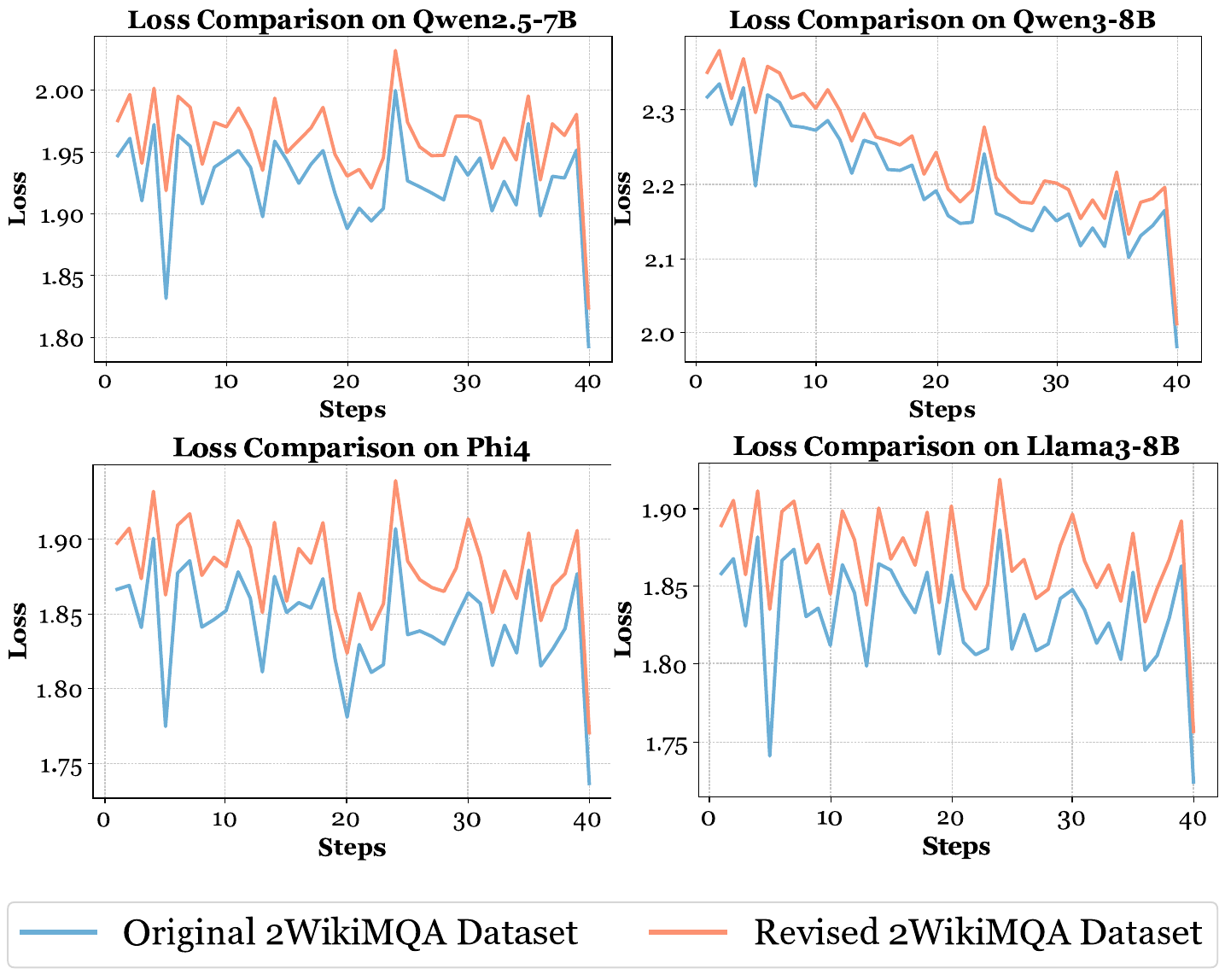}
    \caption{Training Loss Comparison of Language Models on Original and \algname-Revised 2WikiMQA Datasets}
    \label{fig:training}
\end{figure}

To examine whether our dataset defense effectively increases the learning complexity for language models and consequently reduces the likelihood of models memorizing information during the training stage, we conducted an empirical evaluation. Specifically, we fine-tuned four distinct models on both the original 2WikiMQA dataset and our revised dataset\footnote{Details of our fine-tuning settings are provided in Section~\ref{sec:finetune}}. Introducing knowledge conflicts in datasets naturally creates greater complexity during the learning process. Ideally, this discourages models from memorizing knowledge. 

As displayed in Figure~\ref{fig:training}, throughout the fine-tuning process, all models exhibit higher training loss on our revised dataset than on the original dataset, indicating that it exposes them to knowledge conflict situations. The consistently higher training loss values indicate that the revised data poses greater learning difficulty, which in turn mitigates the risk of future data leakage by making it less likely for models to memorize or retain such information.

\subsection{Case Study}

Table~\ref{tab:case_study} illustrates how models may produce correct answers 
without genuinely reasoning over the provided context. In the first setting 
(\emph{Without Context}), the model still predicts the correct year of appointment 
and the correct Olympic Games even though no supporting passages are given. 
This suggests that the model simply recalled memorized knowledge rather than 
performing contextual reasoning. 

In contrast, the second part (\emph{Original/Rewritten Context}) highlights the 
effect of counterfactual rewriting. When the original context explicitly contains 
the correct evidence (\textbf{Marie of Hohenstaufen}), the model outputs the right 
answer. However, after rewriting the critical evidence to state 
\textbf{Joan of Arc}, the model still produces the memorized answer, 
thus failing on the modified case. This discrepancy reveals that the earlier 
success was driven by leakage and memorization rather than true understanding 
of the context. 

\begin{table*}[t]
\centering
\small
\renewcommand{\arraystretch}{1.1} 
\resizebox{\linewidth}{!}{%
\begin{tabular}{lp{10cm}p{2cm}}
\toprule
\textbf{Without Context}: & After Russell D. Moore served at the Southern Baptist Theological Seminary, he became the President of The Ethics \& Religious Liberty Commission (ERLC) in what year? & \bf GPT-4o: 2013 \textcolor{green!80}{$\checkmark$}\\
& Professional cyclist Sara Symington competed in which Olympic Games held in Sydney, Australia? & \bf GPT-4o: 2000 \textcolor{green!80}{$\checkmark$}\\
\hline

\textbf{Original/rewritten Context}: &
\textbf{Question}: Who is the paternal grandmother of Marie of Brabant, Queen of France? 
& \\
& \textbf{Original Context}: ... Marie of Brabant (13 May 1254 – 12 January 1322) was Queen of France from 1274 until 1285 as the second wife of King Philip III. Born in Leuven, Brabant, she was a daughter of Henry III, Duke of Brabant, and Adelaide of Burgundy. ... \par 
Henry III of Brabant (c. 1230 – February 28, 1261, Leuven) was Duke of Brabant between 1248 and his death. \textbf{He was the son of Henry II of Brabant and Marie of Hohenstaufen.} ... 
& \textbf{Marie of Hohenstaufen}\\
& \textbf{Rewritten Context}: ... Marie of Brabant (13 May 1254 – 12 January 1322) was Queen of France from 1274 until 1285 as the second wife of King Philip III. Born in Leuven, Brabant, she was a daughter of Henry III, Duke of Brabant, and Adelaide of Burgundy. ... \par 
Henry III of Brabant (c. 1230 – February 28, 1261, Leuven) was Duke of Brabant between 1248 and his death. \textbf{He was the son of Henry II of \sout{Brabant and Marie of Hohenstaufen.} \colorbox{green!30}{Joan of Arc}} ... 
& \textbf{\sout{Marie of Hohenstaufen} \colorbox{green!30}{Joan of Arc}} \\

\bottomrule
\end{tabular}
}
\caption{Case study showing knowledge leakage and counterfactual defense. 
Top: without any context, the model still gives correct answers (2013, 2000), revealing reliance on memorized knowledge. 
Bottom: in the \textbf{Original Context}, the correct answer (\textbf{Marie of Hohenstaufen}) is present, so the model answers correctly (\checkmark), but this could be due to memorization rather than reasoning. 
In the \textbf{Rewritten Context}, we replace the key evidence with \colorbox{green!30}{Joan of Arc}. The model still outputs the original memorized answer, yielding an error (\texttimes), which confirms that the first success came from recall instead of contextual understanding.}
\label{tab:case_study}
\end{table*}

\section{Related Works}

\subsection{Contamination Detection}

The evaluation of Large Language Models (LLMs) is increasingly complicated by data contamination—the unintended overlap between training data and evaluation benchmarks~\citep{fu2025does, cheng2025survey, xu2024benchmark}. This issue can artificially inflate performance scores, misrepresenting LLMs' true generalization capabilities. Vast pre-training datasets, often web-scraped, frequently include common benchmark samples, leading to deceptively high leaderboard scores that may not reflect real-world robustness~\citep{tonmoy2024comprehensive, xu2024benchmarkdatacontaminationlarge, fu2025does}.

Early detection efforts relied on simple string-matching and overlap detection to find direct textual similarities. However, these methods struggle with nuanced contamination, like paraphrased content or memorized knowledge lacking direct textual overlap~\citep{carlini2021extracting, nasr2023comprehensive}. Consequently, researchers developed advanced techniques that do not require access to proprietary training data~\citep{musawi2025towards, dekoninck2024evading, deng2023investigating}. 
For instance, PaCoST (Paired Confidence Significance Testing) constructs distributionally similar counterparts for benchmark instances and statistically analyzes model confidence on original versus counterpart data to detect significantly higher confidence on originals as potential contamination~\citep{zhang2024pacost, li2023loogle}. Similarly, ConStat adopts a performance-based approach, defining contamination as artificially inflated, non-generalizable performance, and uses statistical analysis to compare model performance on primary and reference benchmarks~\citep{dekoninck2024constat, bommasani2021opportunities}.

Despite these advancements, comprehensively addressing data contamination remains challenging. Notably, specific methodologies for contamination detection in long-context scenarios are lacking in reviewed material~\citep{wang2024leavea, jiang2024longbench, geva2023detecting}. The complexities of extended input sequences may demand novel approaches for identification and mitigation in such contexts~\citep{rajore2024truce, shi2023detecting}.

\subsection{Long Context Benchmark}
Long-context benchmark suites now serve as the standard yardstick for a broad range of evaluations—spanning RAG applications~\citep{cheng2024xragextremecontextcompression,fang2025attentionragattentionguidedcontextpruning,wang2025position}, context-compression methods~\citep{jiang2023llmlinguacompressingpromptsaccelerated,pan2024llmlingua2datadistillationefficient,jiang2024longllmlinguaacceleratingenhancingllms}, and many other long-sequence tasks~\citep{wang2025vrag,zhu2025chain,liu2025shifting,lan2025mcbe,han2024filter,liu2025global,yang2025elaboration,qiu2025latent,zhang2025kabb,zhang2025gam}.
The development of long-context benchmarks like SCROLLS~\citep{shaham2022scrolls} and LongBench~\citep{bai2024longbencha} has been crucial for evaluating LLMs' extended sequence processing. However, these suites are also susceptible to data contamination, where inadvertent inclusion of evaluation samples in training data can inflate performance metrics~\citep{xu2024detectiveqa, oren2023proving}.

To address this, newer contamination-aware benchmarks have been developed, employing various strategies to combat data leakage. Some, like BAMBOO~\citep{dong2024bamboo}, LiveBench~\citep{white2024livebench}, and AcademicEval~\citep{chen2025recent}, feature continuously updated test sets. Others, such as VarBench, introduce dynamic variable perturbation to create more robust and generalizable evaluation scenarios~\citep{qian2024varbench}. A significant limitation with manually updated benchmarks, however, is their heavy reliance on laborious human effort for data collection and maintenance, rendering the process resource-intensive~\citep{sainz2023nlp}. To mitigate this, approaches like Antileak-Bench aim to automatically construct benchmarks from updated real-world knowledge, offering a more scalable solution~\citep{wu2024antileakbench}. Ensuring these updated benchmarks remain truly contamination-free is also challenging, as new data might still contain pre-existing knowledge~\citep{yuan2024lv, lee2024ethic, gao2025u}. Furthermore, undisclosed training data specifics for many LLMs make guaranteeing contamination absence in these benchmarks difficult~\citep{modarressi2025nolima, qi2024long}.

\section{Conclusion}

In this paper, we introduced \algname, a novel framework designed to address the critical challenge of model cheating and knowledge leakage in QA benchmarks. \algname combines perturbation-based detection with counterfactual rewriting to identify and repair leakage points—disrupting memorization while preserving the original evaluative intent of the benchmark. Our experiments on long-context QA benchmarks reveal widespread data leakage across a range of models, including both frontier and smaller open-source LLMs. By applying \algname, we substantially reduce memorization effects and provide a more faithful evaluation of models’ reasoning and generalization capabilities. This work offers a robust and sustainable approach to reinforce benchmark integrity, ensuring that evaluations remain meaningful and resilient against the evolving capabilities of LLMs.

\section*{Limitations} 
\label{sec:limitations}

\algname{} currently targets \emph{with-context} QA and has only been validated on textual long-context benchmarks; applying the same leakage probes and counterfactual rewriting to generation-heavy or multimodal tasks (e.g., summarization, dialogue, code, images) will require new success criteria and may introduce extra computational overhead. Addressing these extensions is left for future work.

\section*{Acknowledgment}
This research is funded by the National Key Research and Development Program of China (Grant No. 2023YFB4503802) and the Natural Science Foundation of Shanghai (Grant No. 25ZR1401175).

\bibliography{custom}

\begin{thebibliography}{72}
\providecommand{\natexlab}[1]{#1}

\bibitem[{Abdin et~al.(2024)Abdin, Aneja, Behl, Bubeck, Eldan, Gunasekar,
  Harrison, Hewett, Javaheripi, Kauffmann, Lee, Lee, Li, Liu, Mendes, Nguyen,
  Price, de~Rosa, Saarikivi, Salim, Shah, Wang, Ward, Wu, Yu, Zhang, and
  Zhang}]{abdin2024phi4technicalreport}
Marah Abdin, Jyoti Aneja, Harkirat Behl, Sébastien Bubeck, Ronen Eldan, Suriya
  Gunasekar, Michael Harrison, Russell~J. Hewett, Mojan Javaheripi, Piero
  Kauffmann, James~R. Lee, Yin~Tat Lee, Yuanzhi Li, Weishung Liu, Caio C.~T.
  Mendes, Anh Nguyen, Eric Price, Gustavo de~Rosa, Olli Saarikivi, Adil Salim,
  Shital Shah, Xin Wang, Rachel Ward, Yue Wu, Dingli Yu, Cyril Zhang, and
  Yi~Zhang. 2024.
\newblock \href {https://arxiv.org/abs/2412.08905} {Phi-4 technical report}.
\newblock \emph{Preprint}, arXiv:2412.08905.

\bibitem[{Bai et~al.(2024{\natexlab{a}})Bai, Lv, Zhang, Lyu, Tang, Huang, Du,
  Liu, Zeng, Hou, Dong, Tang, and Li}]{bai-etal-2024-longbench}
Yushi Bai, Xin Lv, Jiajie Zhang, Hongchang Lyu, Jiankai Tang, Zhidian Huang,
  Zhengxiao Du, Xiao Liu, Aohan Zeng, Lei Hou, Yuxiao Dong, Jie Tang, and
  Juanzi Li. 2024{\natexlab{a}}.
\newblock \href {https://doi.org/10.18653/v1/2024.acl-long.172} {{L}ong{B}ench:
  A bilingual, multitask benchmark for long context understanding}.
\newblock In \emph{Proceedings of the 62nd Annual Meeting of the Association
  for Computational Linguistics (Volume 1: Long Papers)}, pages 3119--3137,
  Bangkok, Thailand. Association for Computational Linguistics.

\bibitem[{Bai et~al.(2024{\natexlab{b}})Bai, Tu, Zhang, Peng, Wang, Lv, Cao,
  Xu, Hou, Dong, Tang, and Li}]{bai2024longbencha}
Yushi Bai, Shangqing Tu, Jiajie Zhang, Hao Peng, Xiaozhi Wang, Xin Lv, Shulin
  Cao, Jiazheng Xu, Lei Hou, Yuxiao Dong, Jie Tang, and Juanzi Li.
  2024{\natexlab{b}}.
\newblock \href {https://arxiv.org/abs/2412.15204} {{{LongBench}} v2: {{Towards
  Deeper Understanding}} and {{Reasoning}} on {{Realistic Long-context
  Multitasks}}}.
\newblock \emph{Preprint}, arXiv:2412.15204.

\bibitem[{Balloccu et~al.(2024)Balloccu, Schmidtová, Lango, and
  Dušek}]{balloccu-etal-2024-leak}
Simone Balloccu, Patrícia Schmidtová, Mateusz Lango, and Ondřej Dušek.
  2024.
\newblock Leak, cheat, repeat: Data contamination and evaluation malpractices
  in closed-source llms.
\newblock In \emph{Proceedings of the 18th Conference of the European Chapter
  of the Association for Computational Linguistics}. Association for
  Computational Linguistics.

\bibitem[{Bommasani et~al.(2021)Bommasani, Hudson, Adeli, Altman, Arora, von
  Arx, Bernstein, Bohg, Bosselut, Brunskill
  et~al.}]{bommasani2021opportunities}
Rishi Bommasani, Drew~A Hudson, Ehsan Adeli, Russ Altman, Simran Arora, Sydney
  von Arx, Michael~S Bernstein, Jeannette Bohg, Antoine Bosselut, Emma
  Brunskill, et~al. 2021.
\newblock On the opportunities and risks of foundation models.
\newblock \emph{arXiv preprint arXiv:2108.07258}.

\bibitem[{Carlini et~al.(2021)Carlini, Tramer, Wallace, Jagielski, Roberts,
  Bowman, Hardt, Papernot et~al.}]{carlini2021extracting}
Nicholas Carlini, Florian Tramer, Eric Wallace, Miroslav Jagielski, Matthew
  Roberts, Samuel~R Bowman, Moritz Hardt, Nicolas Papernot, et~al. 2021.
\newblock \href
  {https://www.usenix.org/conference/usenixsecurity21/presentation/carlini-extracting}
  {Extracting training data from large language models}.
\newblock In \emph{30th USENIX Security Symposium (USENIX Security 21)}, pages
  263--280.

\bibitem[{Chen et~al.(2025)Chen, Chen, Li, Jiang, Wan, He, Ran, Gu, Li, Xie,
  and Ray}]{chen2025recent}
Simin Chen, Yiming Chen, Zexin Li, Yifan Jiang, Zhongwei Wan, Yixin He, Dezhi
  Ran, Tianle Gu, Haizhou Li, Tao Xie, and Baishakhi Ray. 2025.
\newblock \href {https://doi.org/10.48550/arXiv.2502.17521} {Recent
  {{Advances}} in {{Large Langauge Model Benchmarks}} against {{Data
  Contamination}}: {{From Static}} to {{Dynamic Evaluation}}}.
\newblock \emph{Preprint}, arXiv:2502.17521.

\bibitem[{Cheng et~al.(2024)Cheng, Wang, Zhang, Ge, Chen, Wei, Zhang, and
  Zhao}]{cheng2024xragextremecontextcompression}
Xin Cheng, Xun Wang, Xingxing Zhang, Tao Ge, Si-Qing Chen, Furu Wei, Huishuai
  Zhang, and Dongyan Zhao. 2024.
\newblock \href {https://arxiv.org/abs/2405.13792} {xrag: Extreme context
  compression for retrieval-augmented generation with one token}.
\newblock \emph{Preprint}, arXiv:2405.13792.

\bibitem[{Cheng et~al.(2025)Cheng, Chang, and Wu}]{cheng2025survey}
Yuxing Cheng, Yi~Chang, and Yuan Wu. 2025.
\newblock \href {https://doi.org/10.48550/arXiv.2502.14425} {A {{Survey}} on
  {{Data Contamination}} for {{Large Language Models}}}.
\newblock \emph{Preprint}, arXiv:2502.14425.

\bibitem[{Daniel~Han and team(2023)}]{unsloth}
Michael~Han Daniel~Han and Unsloth team. 2023.
\newblock \href {http://github.com/unslothai/unsloth} {Unsloth}.

\bibitem[{DeepSeek-AI(2025)}]{deepseekai2025deepseekr1incentivizingreasoningcapability}
DeepSeek-AI. 2025.
\newblock \href {https://arxiv.org/abs/2501.12948} {Deepseek-r1: Incentivizing
  reasoning capability in llms via reinforcement learning}.
\newblock \emph{Preprint}, arXiv:2501.12948.

\bibitem[{Dekoninck et~al.(2024{\natexlab{a}})Dekoninck, Mueller, and
  Vechev}]{dekoninck2024constat}
Jasper Dekoninck, Mark~Niklas Mueller, and Martin Vechev. 2024{\natexlab{a}}.
\newblock {{ConStat}}: {{Performance-Based Contamination Detection}} in {{Large
  Language Models}}.
\newblock In \emph{The {{Thirty-eighth Annual Conference}} on {{Neural
  Information Processing Systems}}}.

\bibitem[{Dekoninck et~al.(2024{\natexlab{b}})Dekoninck, M{\"u}ller, Baader,
  Fischer, and Vechev}]{dekoninck2024evading}
Jasper Dekoninck, Mark~Niklas M{\"u}ller, Maximilian Baader, Marc Fischer, and
  Martin Vechev. 2024{\natexlab{b}}.
\newblock Evading data contamination detection for language models is (too)
  easy.
\newblock \emph{arXiv preprint arXiv:2402.02823}.

\bibitem[{Deng et~al.(2023)Deng, Zhao, Tang, Gerstein, and
  Cohan}]{deng2023investigating}
Chunyuan Deng, Yilun Zhao, Xiangru Tang, Mark Gerstein, and Arman Cohan. 2023.
\newblock Investigating data contamination in modern benchmarks for large
  language models.
\newblock \emph{arXiv preprint arXiv:2311.09783}.

\bibitem[{Deng et~al.(2024)Deng, Zhao, Tang, Gerstein, and
  Cohan}]{deng2024benchmark}
Chunyuan Deng, Yilun Zhao, Xiangru Tang, Mark Gerstein, and Arman Cohan. 2024.
\newblock \href {https://openreview.net/forum?id=a34bgvner1} {Benchmark
  probing: Investigating data leakage in large language models}.
\newblock In \emph{NeurIPS 2023 Workshop on Backdoors in Deep Learning - The
  Good, the Bad, and the Ugly}.

\bibitem[{Dong et~al.(2024)Dong, Tang, Li, Zhao, and Wen}]{dong2024bamboo}
Zican Dong, Tianyi Tang, Junyi Li, Wayne~Xin Zhao, and Ji-Rong Wen. 2024.
\newblock {{BAMBOO}}: {{A Comprehensive Benchmark}} for {{Evaluating Long Text
  Modeling Capacities}} of {{Large Language Models}}.
\newblock In \emph{Proceedings of the 2024 {{Joint International Conference}}
  on {{Computational Linguistics}}, {{Language Resources}} and {{Evaluation}}
  ({{LREC-COLING}} 2024)}, pages 2086--2099, Torino, Italia. {ELRA and ICCL}.

\bibitem[{et~al(2024)}]{grattafiori2024llama3herdmodels}
Aaron~Grattafiori et~al. 2024.
\newblock \href {https://arxiv.org/abs/2407.21783} {The llama 3 herd of
  models}.
\newblock \emph{Preprint}, arXiv:2407.21783.

\bibitem[{Fang et~al.(2025)Fang, Sun, Shi, and
  Gu}]{fang2025attentionragattentionguidedcontextpruning}
Yixiong Fang, Tianran Sun, Yuling Shi, and Xiaodong Gu. 2025.
\newblock \href {https://arxiv.org/abs/2503.10720} {Attentionrag:
  Attention-guided context pruning in retrieval-augmented generation}.
\newblock \emph{Preprint}, arXiv:2503.10720.

\bibitem[{Fu et~al.(2025)Fu, Uzuner, Yetisgen, and Xia}]{fu2025does}
Yujuan Fu, Ozlem Uzuner, Meliha Yetisgen, and Fei Xia. 2025.
\newblock Does {{Data Contamination Detection Work}} ({{Well}}) for {{LLMs}}?
  {{A Survey}} and {{Evaluation}} on {{Detection Assumptions}}.
\newblock In \emph{Findings of the {{Association}} for {{Computational
  Linguistics}}: {{NAACL}} 2025}, pages 5235--5256, Albuquerque, New Mexico.
  Association for Computational Linguistics.

\bibitem[{Gao et~al.(2025)Gao, Xiong, Wu, Huang, Li, and Wang}]{gao2025u}
Yunfan Gao, Yun Xiong, Wenlong Wu, Zijing Huang, Bohan Li, and Haofen Wang.
  2025.
\newblock U-niah: Unified rag and llm evaluation for long context
  needle-in-a-haystack.
\newblock \emph{arXiv preprint arXiv:2503.00353}.

\bibitem[{Geva et~al.(2023)Geva, Schuster, Berant, and
  Levy}]{geva2023detecting}
Mor Geva, Tal Schuster, Jonathan Berant, and Omer Levy. 2023.
\newblock \href {https://aclanthology.org/2023.emnlp-main.80} {Detecting
  pretraining data from large language models}.
\newblock In \emph{Proceedings of the 2023 Conference on Empirical Methods in
  Natural Language Processing}, pages 1207--1220. Association for Computational
  Linguistics.

\bibitem[{Golchin and Surdeanu(2025)}]{golchin2025datacontaminationquiztool}
Shahriar Golchin and Mihai Surdeanu. 2025.
\newblock \href {https://arxiv.org/abs/2311.06233} {Data contamination quiz: A
  tool to detect and estimate contamination in large language models}.
\newblock \emph{Preprint}, arXiv:2311.06233.

\bibitem[{Han et~al.(2024)Han, Liu, Zhang, Ding, Wang, Chen, Yan, and
  Huang}]{han2024filter}
Yuhang Han, Xuyang Liu, Zihan Zhang, Pengxiang Ding, Donglin Wang, Honggang
  Chen, Qingsen Yan, and Siteng Huang. 2024.
\newblock Filter, correlate, compress: Training-free token reduction for mllm
  acceleration.
\newblock \emph{arXiv preprint arXiv:2411.17686}.

\bibitem[{Ho et~al.(2020)Ho, Duong~Nguyen, Sugawara, and
  Aizawa}]{ho-etal-2020-constructing}
Xanh Ho, Anh-Khoa Duong~Nguyen, Saku Sugawara, and Akiko Aizawa. 2020.
\newblock \href {https://doi.org/10.18653/v1/2020.coling-main.580}
  {Constructing a multi-hop {QA} dataset for comprehensive evaluation of
  reasoning steps}.
\newblock In \emph{Proceedings of the 28th International Conference on
  Computational Linguistics}, pages 6609--6625, Barcelona, Spain (Online).
  International Committee on Computational Linguistics.

\bibitem[{Hu et~al.(2021)Hu, Shen, Wallis, Allen-Zhu, Li, Wang, Wang, and
  Chen}]{hu2021loralowrankadaptationlarge}
Edward~J. Hu, Yelong Shen, Phillip Wallis, Zeyuan Allen-Zhu, Yuanzhi Li, Shean
  Wang, Lu~Wang, and Weizhu Chen. 2021.
\newblock \href {https://arxiv.org/abs/2106.09685} {Lora: Low-rank adaptation
  of large language models}.
\newblock \emph{Preprint}, arXiv:2106.09685.

\bibitem[{Hurst et~al.(2024)Hurst, Lerer, Goucher, Perelman, Ramesh, Clark,
  Ostrow, Welihinda, Hayes, Radford et~al.}]{hurst2024gpt}
Aaron Hurst, Adam Lerer, Adam~P Goucher, Adam Perelman, Aditya Ramesh, Aidan
  Clark, AJ~Ostrow, Akila Welihinda, Alan Hayes, Alec Radford, et~al. 2024.
\newblock Gpt-4o system card.
\newblock \emph{arXiv preprint arXiv:2410.21276}.

\bibitem[{Jain et~al.(2024)Jain, Han, Gu, Li, Yan, Zhang, Wang, Solar-Lezama,
  Sen, and Stoica}]{jain2024livecodebenchholisticcontaminationfree}
Naman Jain, King Han, Alex Gu, Wen-Ding Li, Fanjia Yan, Tianjun Zhang, Sida
  Wang, Armando Solar-Lezama, Koushik Sen, and Ion Stoica. 2024.
\newblock \href {https://arxiv.org/abs/2403.07974} {Livecodebench: Holistic and
  contamination free evaluation of large language models for code}.
\newblock \emph{Preprint}, arXiv:2403.07974.

\bibitem[{Jiang et~al.(2024{\natexlab{a}})Jiang, Wei, Pan, Li, Liang, Tay, Wei,
  Yan, Wang, Li et~al.}]{jiang2024longbench}
Hao Jiang, Nan Wei, Xiao Pan, Yu~Li, Xiaochen Liang, Yi~Tay, Jilan Wei, Xiaohui
  Yan, Xiaozhi Wang, Juanzi Li, et~al. 2024{\natexlab{a}}.
\newblock \href {https://openreview.net/forum?id=yYF4mqwR9y} {{LongBench}: A
  bilingual, multitask benchmark for long context understanding}.
\newblock \emph{Transactions on Machine Learning Research}.

\bibitem[{Jiang et~al.(2023)Jiang, Wu, Lin, Yang, and
  Qiu}]{jiang2023llmlinguacompressingpromptsaccelerated}
Huiqiang Jiang, Qianhui Wu, Chin-Yew Lin, Yuqing Yang, and Lili Qiu. 2023.
\newblock \href {https://arxiv.org/abs/2310.05736} {Llmlingua: Compressing
  prompts for accelerated inference of large language models}.
\newblock \emph{Preprint}, arXiv:2310.05736.

\bibitem[{Jiang et~al.(2024{\natexlab{b}})Jiang, Wu, Luo, Li, Lin, Yang, and
  Qiu}]{jiang2024longllmlinguaacceleratingenhancingllms}
Huiqiang Jiang, Qianhui Wu, Xufang Luo, Dongsheng Li, Chin-Yew Lin, Yuqing
  Yang, and Lili Qiu. 2024{\natexlab{b}}.
\newblock \href {https://arxiv.org/abs/2310.06839} {Longllmlingua: Accelerating
  and enhancing llms in long context scenarios via prompt compression}.
\newblock \emph{Preprint}, arXiv:2310.06839.

\bibitem[{Jiang et~al.(2025)Jiang, Lin, Cao, Tian, Kang, Wang, Sun, and
  Han}]{jiang2025deepretrievalhackingrealsearch}
Pengcheng Jiang, Jiacheng Lin, Lang Cao, Runchu Tian, SeongKu Kang, Zifeng
  Wang, Jimeng Sun, and Jiawei Han. 2025.
\newblock \href {https://arxiv.org/abs/2503.00223} {Deepretrieval: Hacking real
  search engines and retrievers with large language models via reinforcement
  learning}.
\newblock \emph{Preprint}, arXiv:2503.00223.

\bibitem[{Lan et~al.(2025)Lan, Su, Liu, Wang, Chang, Li, and Gao}]{lan2025mcbe}
Tian Lan, Xiangdong Su, Xu~Liu, Ruirui Wang, Ke~Chang, Jiang Li, and Guanglai
  Gao. 2025.
\newblock Mcbe: A multi-task chinese bias evaluation benchmark for large
  language models.
\newblock \emph{arXiv preprint arXiv:2507.02088}.

\bibitem[{Lee et~al.(2024)Lee, Yoon, Jang, Lee, Song, Kim, and
  Kang}]{lee2024ethic}
Taewhoo Lee, Chanwoong Yoon, Kyochul Jang, Donghyeon Lee, Minju Song, Hyunjae
  Kim, and Jaewoo Kang. 2024.
\newblock Ethic: Evaluating large language models on long-context tasks with
  high information coverage.
\newblock \emph{arXiv preprint arXiv:2410.16848}.

\bibitem[{Li et~al.(2023)Li, Wang, Zheng, and Zhang}]{li2023loogle}
Jiaqi Li, Mengmeng Wang, Zilong Zheng, and Muhan Zhang. 2023.
\newblock Loogle: Can long-context language models understand long contexts?
\newblock \emph{arXiv preprint arXiv:2311.04939}.

\bibitem[{Liu et~al.(2024{\natexlab{a}})Liu, Feng, Xue, Wang, Wu, Lu, Zhao,
  Deng, Zhang, Ruan et~al.}]{liu2024deepseek}
Aixin Liu, Bei Feng, Bing Xue, Bingxuan Wang, Bochao Wu, Chengda Lu, Chenggang
  Zhao, Chengqi Deng, Chenyu Zhang, Chong Ruan, et~al. 2024{\natexlab{a}}.
\newblock Deepseek-v3 technical report.
\newblock \emph{arXiv preprint arXiv:2412.19437}.

\bibitem[{Liu et~al.(2024{\natexlab{b}})Liu, Chen, Ji, Zhou, Chen, and
  Wang}]{liu2024rag}
Wanlong Liu, Junying Chen, Ke~Ji, Li~Zhou, Wenyu Chen, and Benyou Wang.
  2024{\natexlab{b}}.
\newblock Rag-instruct: Boosting llms with diverse retrieval-augmented
  instructions.
\newblock \emph{arXiv preprint arXiv:2501.00353}.

\bibitem[{Liu et~al.(2025{\natexlab{a}})Liu, Wang, Han, Wang, Yuan, Song,
  Zheng, Zhang, Huang, and Chen}]{liu2025global}
Xuyang Liu, Ziming Wang, Yuhang Han, Yingyao Wang, Jiale Yuan, Jun Song,
  Bo~Zheng, Linfeng Zhang, Siteng Huang, and Honggang Chen. 2025{\natexlab{a}}.
\newblock Global compression commander: Plug-and-play inference acceleration
  for high-resolution large vision-language models.
\newblock \emph{arXiv preprint arXiv:2501.05179}.

\bibitem[{Liu et~al.(2025{\natexlab{b}})Liu, Wen, Wang, Chen, Tao, Wang, Jin,
  Zou, Wang, Liao et~al.}]{liu2025shifting}
Xuyang Liu, Zichen Wen, Shaobo Wang, Junjie Chen, Zhishan Tao, Yubo Wang,
  Xiangqi Jin, Chang Zou, Yiyu Wang, Chenfei Liao, et~al. 2025{\natexlab{b}}.
\newblock Shifting ai efficiency from model-centric to data-centric
  compression.
\newblock \emph{arXiv preprint arXiv:2505.19147}.

\bibitem[{Modarressi et~al.(2025)Modarressi, Deilamsalehy, Dernoncourt, Bui,
  Rossi, Yoon, and Sch{\"u}tze}]{modarressi2025nolima}
Ali Modarressi, Hanieh Deilamsalehy, Franck Dernoncourt, Trung Bui, Ryan~A
  Rossi, Seunghyun Yoon, and Hinrich Sch{\"u}tze. 2025.
\newblock Nolima: Long-context evaluation beyond literal matching.
\newblock \emph{arXiv preprint arXiv:2502.05167}.

\bibitem[{Musawi and Lu(2025)}]{musawi2025towards}
Rahmatullah Musawi and Sheng Lu. 2025.
\newblock \href {https://arxiv.org/abs/2505.08389} {Towards contamination
  resistant benchmarks}.
\newblock \emph{arXiv preprint arXiv:2505.08389}.

\bibitem[{Nasr et~al.(2023)Nasr, Carlini, Roberts, Zhang, Song, Erlingsson,
  Papernot, Lee, and Raffel}]{nasr2023comprehensive}
Milad Nasr, Nicholas Carlini, Matthew Roberts, Tianyi Zhang, Katherine Song,
  {\'U}lfar Erlingsson, Nicolas Papernot, Chris Lee, and Colin Raffel. 2023.
\newblock \href {https://arxiv.org/abs/2307.06244} {A comprehensive study of
  memorization in large language models}.
\newblock \emph{arXiv preprint arXiv:2307.06244}.

\bibitem[{Oren et~al.(2023)Oren, Meister, Chatterji, Ladhak, and
  Hashimoto}]{oren2023proving}
Yonatan Oren, Nicole Meister, Niladri~S Chatterji, Faisal Ladhak, and Tatsunori
  Hashimoto. 2023.
\newblock Proving test set contamination in black-box language models.
\newblock In \emph{The Twelfth International Conference on Learning
  Representations}.

\bibitem[{Pan et~al.(2024)Pan, Wu, Jiang, Xia, Luo, Zhang, Lin, Rühle, Yang,
  Lin, Zhao, Qiu, and Zhang}]{pan2024llmlingua2datadistillationefficient}
Zhuoshi Pan, Qianhui Wu, Huiqiang Jiang, Menglin Xia, Xufang Luo, Jue Zhang,
  Qingwei Lin, Victor Rühle, Yuqing Yang, Chin-Yew Lin, H.~Vicky Zhao, Lili
  Qiu, and Dongmei Zhang. 2024.
\newblock \href {https://arxiv.org/abs/2403.12968} {Llmlingua-2: Data
  distillation for efficient and faithful task-agnostic prompt compression}.
\newblock \emph{Preprint}, arXiv:2403.12968.

\bibitem[{Qi et~al.(2024)Qi, Xu, Guo, Wang, Zhang, and Xu}]{qi2024long}
Zehan Qi, Rongwu Xu, Zhijiang Guo, Cunxiang Wang, Hao Zhang, and Wei Xu. 2024.
\newblock Long\textsuperscript{2} rag: Evaluating long-context \& long-form
  retrieval-augmented generation with key point recall.
\newblock \emph{arXiv preprint arXiv:2410.23000}.

\bibitem[{Qian et~al.(2025)Qian, Liu, Zhang, Mao, Lian, Dou, and
  Huang}]{qian2025memoragboostinglongcontext}
Hongjin Qian, Zheng Liu, Peitian Zhang, Kelong Mao, Defu Lian, Zhicheng Dou,
  and Tiejun Huang. 2025.
\newblock \href {https://arxiv.org/abs/2409.05591} {Memorag: Boosting long
  context processing with global memory-enhanced retrieval augmentation}.
\newblock \emph{Preprint}, arXiv:2409.05591.

\bibitem[{Qian et~al.(2024)Qian, Wan, Tang, Wang, Zhang, Chen, and
  Yu}]{qian2024varbench}
Kun Qian, Shunji Wan, Claudia Tang, Youzhi Wang, Xuanming Zhang, Maximillian
  Chen, and Zhou Yu. 2024.
\newblock Varbench: Robust language model benchmarking through dynamic variable
  perturbation.
\newblock In \emph{Findings of the Association for Computational Linguistics:
  EMNLP 2024}, pages 16131--16161.

\bibitem[{Qiu et~al.(2025)Qiu, Shi, Zhao, Zhu, Zhang, and Feng}]{qiu2025latent}
Yilun Qiu, Tianhao Shi, Xiaoyan Zhao, Fengbin Zhu, Yang Zhang, and Fuli Feng.
  2025.
\newblock Latent inter-user difference modeling for llm personalization.
\newblock \emph{arXiv preprint arXiv:2507.20849}.

\bibitem[{Qwen(2025)}]{qwen2025qwen25technicalreport}
Qwen. 2025.
\newblock \href {https://arxiv.org/abs/2412.15115} {Qwen2.5 technical report}.
\newblock \emph{Preprint}, arXiv:2412.15115.

\bibitem[{Rajore et~al.(2024)Rajore, Chandran, Sitaram, Gupta, Sharma, Mittal,
  and Swaminathan}]{rajore2024truce}
Tanmay Rajore, Nishanth Chandran, Sunayana Sitaram, Divya Gupta, Rahul Sharma,
  Kashish Mittal, and Manohar Swaminathan. 2024.
\newblock Truce: Private benchmarking to prevent contamination and improve
  comparative evaluation of llms.
\newblock \emph{arXiv preprint arXiv:2403.00393}.

\bibitem[{Sainz et~al.(2023{\natexlab{a}})Sainz, Campos, Garc{\'i}a-Ferrero,
  Etxaniz, de~Lacalle, and Agirre}]{sainz-etal-2023-nlp}
Oscar Sainz, Jon Campos, Iker Garc{\'i}a-Ferrero, Julen Etxaniz, Oier~Lopez
  de~Lacalle, and Eneko Agirre. 2023{\natexlab{a}}.
\newblock \href {https://doi.org/10.18653/v1/2023.findings-emnlp.722} {{NLP}
  evaluation in trouble: On the need to measure {LLM} data contamination for
  each benchmark}.
\newblock In \emph{Findings of the Association for Computational Linguistics:
  EMNLP 2023}, pages 10776--10787, Singapore. Association for Computational
  Linguistics.

\bibitem[{Sainz et~al.(2023{\natexlab{b}})Sainz, Campos, Garc{\'\i}a-Ferrero,
  Etxaniz, de~Lacalle, and Agirre}]{sainz2023nlp}
Oscar Sainz, Jon~Ander Campos, Iker Garc{\'\i}a-Ferrero, Julen Etxaniz,
  Oier~Lopez de~Lacalle, and Eneko Agirre. 2023{\natexlab{b}}.
\newblock Nlp evaluation in trouble: On the need to measure llm data
  contamination for each benchmark.
\newblock \emph{arXiv preprint arXiv:2310.18018}.

\bibitem[{Shaham et~al.(2022)Shaham, Segal, Ivgi, Efrat, Yoran, Haviv, Gupta,
  Xiong, Geva, Berant, and Levy}]{shaham2022scrolls}
Uri Shaham, Elad Segal, Maor Ivgi, Avia Efrat, Ori Yoran, Adi Haviv, Ankit
  Gupta, Wenhan Xiong, Mor Geva, Jonathan Berant, and Omer Levy. 2022.
\newblock \href {https://doi.org/10.18653/v1/2022.emnlp-main.823} {{{SCROLLS}}:
  {{Standardized CompaRison Over Long Language Sequences}}}.
\newblock In \emph{Proceedings of the 2022 {{Conference}} on {{Empirical
  Methods}} in {{Natural Language Processing}}}, pages 12007--12021, Abu Dhabi,
  United Arab Emirates. Association for Computational Linguistics.

\bibitem[{Shi et~al.(2023)Shi, Ajith, Xia, Huang, Liu, Blevins, Chen, and
  Zettlemoyer}]{shi2023detecting}
Weijia Shi, Anirudh Ajith, Mengzhou Xia, Yangsibo Huang, Daogao Liu, Terra
  Blevins, Danqi Chen, and Luke Zettlemoyer. 2023.
\newblock Detecting pretraining data from large language models.
\newblock \emph{arXiv preprint arXiv:2310.16789}.

\bibitem[{Tonmoy et~al.(2024)Tonmoy, Zaman, Jain, Rani, Rawte, Chadha, and
  Das}]{tonmoy2024comprehensive}
SM~Tonmoy, SM~Zaman, Vinija Jain, Anku Rani, Vipula Rawte, Aman Chadha, and
  Amitava Das. 2024.
\newblock A comprehensive survey of hallucination mitigation techniques in
  large language models.
\newblock \emph{arXiv preprint arXiv:2401.01313}, 6.

\bibitem[{Trivedi et~al.(2022)Trivedi, Balasubramanian, Khot, and
  Sabharwal}]{trivedi-etal-2022-musique}
Harsh Trivedi, Niranjan Balasubramanian, Tushar Khot, and Ashish Sabharwal.
  2022.
\newblock \href {https://doi.org/10.1162/tacl_a_00475} {{M}u{S}i{Q}ue: Multihop
  questions via single-hop question composition}.
\newblock \emph{Transactions of the Association for Computational Linguistics},
  10:539--554.

\bibitem[{Wang et~al.(2024)Wang, Chen, Cheng, Liao, Zhang, Wu, Yu, Xu, Zhang,
  Luo, Li, Yang, Huang, and Li}]{wang2024leavea}
Minzheng Wang, Longze Chen, Fu~Cheng, Shengyi Liao, Xinghua Zhang, Bingli Wu,
  Haiyang Yu, Nan Xu, Lei Zhang, Run Luo, Yunshui Li, Min Yang, Fei Huang, and
  Yongbin Li. 2024.
\newblock \href {https://doi.org/10.18653/v1/2024.emnlp-main.322} {Leave {{No
  Document Behind}}: {{Benchmarking Long-Context LLMs}} with {{Extended
  Multi-Doc QA}}}.
\newblock In \emph{Proceedings of the 2024 {{Conference}} on {{Empirical
  Methods}} in {{Natural Language Processing}}}, pages 5627--5646, Miami,
  Florida, USA. Association for Computational Linguistics.

\bibitem[{Wang et~al.(2025{\natexlab{a}})Wang, Ding, Chen, Wu, Wang, Xie, and
  Zhao}]{wang2025vidorag}
Qiuchen Wang, Ruixue Ding, Zehui Chen, Weiqi Wu, Shihang Wang, Pengjun Xie, and
  Feng Zhao. 2025{\natexlab{a}}.
\newblock Vidorag: Visual document retrieval-augmented generation via dynamic
  iterative reasoning agents.
\newblock \emph{arXiv preprint arXiv:2502.18017}.

\bibitem[{Wang et~al.(2025{\natexlab{b}})Wang, Ding, Zeng, Chen, Chen, Wang,
  Xie, Huang, and Zhao}]{wang2025vrag}
Qiuchen Wang, Ruixue Ding, Yu~Zeng, Zehui Chen, Lin Chen, Shihang Wang, Pengjun
  Xie, Fei Huang, and Feng Zhao. 2025{\natexlab{b}}.
\newblock Vrag-rl: Empower vision-perception-based rag for visually rich
  information understanding via iterative reasoning with reinforcement
  learning.
\newblock \emph{arXiv preprint arXiv:2505.22019}.

\bibitem[{Wang et~al.(2025{\natexlab{c}})Wang, Xiong, Wang, Li, Chu, and
  Zeng}]{wang2025position}
Yifei Wang, Feng Xiong, Yong Wang, Linjing Li, Xiangxiang Chu, and Daniel~Dajun
  Zeng. 2025{\natexlab{c}}.
\newblock Position bias mitigates position bias: Mitigate position bias through
  inter-position knowledge distillation.
\newblock \emph{arXiv preprint arXiv:2508.15709}.

\bibitem[{White et~al.(2024)White, Dooley, Roberts, Pal, Feuer, Jain,
  {Shwartz-Ziv}, Jain, Saifullah, Dey, {Shubh-Agrawal}, Sandha, Naidu, Hegde,
  LeCun, Goldstein, Neiswanger, and Goldblum}]{white2024livebench}
Colin White, Samuel Dooley, Manley Roberts, Arka Pal, Benjamin Feuer,
  Siddhartha Jain, Ravid {Shwartz-Ziv}, Neel Jain, Khalid Saifullah, Sreemanti
  Dey, {Shubh-Agrawal}, Sandeep~Singh Sandha, Siddartha~Venkat Naidu, Chinmay
  Hegde, Yann LeCun, Tom Goldstein, Willie Neiswanger, and Micah Goldblum.
  2024.
\newblock {{LiveBench}}: {{A Challenging}}, {{Contamination-Limited LLM
  Benchmark}}.
\newblock In \emph{The {{Thirteenth International Conference}} on {{Learning
  Representations}}}.

\bibitem[{Wu et~al.(2024)Wu, Pan, Xie, Zhou, Zhao, Ma, Du, Mao, Luu, and
  Wang}]{wu2024antileakbench}
Xiaobao Wu, Liangming Pan, Yuxi Xie, Ruiwen Zhou, Shuai Zhao, Yubo Ma, Mingzhe
  Du, Rui Mao, Anh~Tuan Luu, and William~Yang Wang. 2024.
\newblock Antileakbench: Preventing data contamination by automatically
  constructing benchmarks with updated real-world knowledge.
\newblock \emph{arXiv preprint arXiv:2412.13670}.

\bibitem[{Xu et~al.(2024{\natexlab{a}})Xu, Guan, Greene, and
  Kechadi}]{xu2024benchmark}
Cheng Xu, Shuhao Guan, Derek Greene, and M.-Tahar Kechadi. 2024{\natexlab{a}}.
\newblock \href {https://doi.org/10.48550/arXiv.2406.04244} {Benchmark {{Data
  Contamination}} of {{Large Language Models}}: {{A Survey}}}.
\newblock \emph{Preprint}, arXiv:2406.04244.

\bibitem[{Xu et~al.(2024{\natexlab{b}})Xu, Guan, Greene, and
  Kechadi}]{xu2024benchmarkdatacontaminationlarge}
Cheng Xu, Shuhao Guan, Derek Greene, and M-Tahar Kechadi. 2024{\natexlab{b}}.
\newblock \href {https://arxiv.org/abs/2406.04244} {Benchmark data
  contamination of large language models: A survey}.
\newblock \emph{Preprint}, arXiv:2406.04244.

\bibitem[{Xu et~al.(2024{\natexlab{c}})Xu, Wang, Fan, and
  Liu}]{xu2024benchmarkingbenchmarkleakagelarge}
Ruijie Xu, Zengzhi Wang, Run-Ze Fan, and Pengfei Liu. 2024{\natexlab{c}}.
\newblock \href {https://arxiv.org/abs/2404.18824} {Benchmarking benchmark
  leakage in large language models}.
\newblock \emph{Preprint}, arXiv:2404.18824.

\bibitem[{Xu et~al.(2024{\natexlab{d}})Xu, Ye, Liu, Liu, Sun, Liu, Guo, Li,
  Liu, Huang et~al.}]{xu2024detectiveqa}
Zhe Xu, Jiasheng Ye, Xiaoran Liu, Xiangyang Liu, Tianxiang Sun, Zhigeng Liu,
  Qipeng Guo, Linlin Li, Qun Liu, Xuanjing Huang, et~al. 2024{\natexlab{d}}.
\newblock Detectiveqa: Evaluating long-context reasoning on detective novels.
\newblock \emph{arXiv preprint arXiv:2409.02465}.

\bibitem[{Yang et~al.(2025)Yang, Liu, Huang, Zhang, Zhang, Zhang, and
  Lei}]{yang2025elaboration}
Xinwei Yang, Zhaofeng Liu, Chen Huang, Jiashuai Zhang, Tong Zhang, Yifan Zhang,
  and Wenqiang Lei. 2025.
\newblock Elaboration: A comprehensive benchmark on human-llm competitive
  programming.
\newblock \emph{arXiv preprint arXiv:2505.16667}.

\bibitem[{Yang et~al.(2018)Yang, Qi, Zhang, Bengio, Cohen, Salakhutdinov, and
  Manning}]{yang-etal-2018-hotpotqa}
Zhilin Yang, Peng Qi, Saizheng Zhang, Yoshua Bengio, William Cohen, Ruslan
  Salakhutdinov, and Christopher~D. Manning. 2018.
\newblock \href {https://doi.org/10.18653/v1/D18-1259} {{H}otpot{QA}: A dataset
  for diverse, explainable multi-hop question answering}.
\newblock In \emph{Proceedings of the 2018 Conference on Empirical Methods in
  Natural Language Processing}, pages 2369--2380, Brussels, Belgium.
  Association for Computational Linguistics.

\bibitem[{Yuan et~al.(2024)Yuan, Ning, Zhou, Yang, Li, Zhuang, Tan, Yao, Lin,
  Li et~al.}]{yuan2024lv}
Tao Yuan, Xuefei Ning, Dong Zhou, Zhijie Yang, Shiyao Li, Minghui Zhuang,
  Zheyue Tan, Zhuyu Yao, Dahua Lin, Boxun Li, et~al. 2024.
\newblock Lv-eval: A balanced long-context benchmark with 5 length levels up to
  256k.
\newblock \emph{arXiv preprint arXiv:2402.05136}.

\bibitem[{Zhang et~al.(2024)Zhang, Lin, and Wan}]{zhang2024pacost}
Huixuan Zhang, Yun Lin, and Xiaojun Wan. 2024.
\newblock {{PaCoST}}: {{Paired Confidence Significance Testing}} for
  {{Benchmark Contamination Detection}} in {{Large Language Models}}.
\newblock In \emph{Findings of the {{Association}} for {{Computational
  Linguistics}}: {{EMNLP}} 2024}, pages 1794--1809.

\bibitem[{Zhang et~al.(2025{\natexlab{a}})Zhang, Fan, Lin, Chen, Jiang, Chai,
  Wang, and Wang}]{zhang2025gam}
Jusheng Zhang, Yijia Fan, Wenjun Lin, Ruiqi Chen, Haoyi Jiang, Wenhao Chai,
  Jian Wang, and Keze Wang. 2025{\natexlab{a}}.
\newblock Gam-agent: Game-theoretic and uncertainty-aware collaboration for
  complex visual reasoning.
\newblock \emph{arXiv preprint arXiv:2505.23399}.

\bibitem[{Zhang et~al.(2025{\natexlab{b}})Zhang, Huang, Fan, Liu, Li, Yang,
  Yao, Wang, and Wang}]{zhang2025kabb}
Jusheng Zhang, Zimeng Huang, Yijia Fan, Ningyuan Liu, Mingyan Li, Zhuojie Yang,
  Jiawei Yao, Jian Wang, and Keze Wang. 2025{\natexlab{b}}.
\newblock \href {https://openreview.net/forum?id=AKvy9a4jho} {{KABB}:
  Knowledge-aware bayesian bandits for dynamic expert coordination in
  multi-agent systems}.
\newblock In \emph{Forty-second International Conference on Machine Learning}.

\bibitem[{Zhu et~al.(2025)Zhu, Wei, Zhao, Wu, Zou, Ran, Wang, Sun, Zhang, and
  Li}]{zhu2025chain}
Dawei Zhu, Xiyu Wei, Guangxiang Zhao, Wenhao Wu, Haosheng Zou, Junfeng Ran, Xun
  Wang, Lin Sun, Xiangzheng Zhang, and Sujian Li. 2025.
\newblock Chain-of-thought matters: improving long-context language models with
  reasoning path supervision.
\newblock \emph{arXiv preprint arXiv:2502.20790}.

\end{thebibliography}
\appendix



\section{Finetune Settings}\label{sec:finetune}
We fine-tune four models on both datasets using one NVIDIA A100-SXM4-80GB GPU for 10 epochs. The fine-tuning process leverages the Unsloth library~\cite{unsloth} and incorporates LoRA~\cite{hu2021loralowrankadaptationlarge} with a rank of 32 and an alpha of 32. The initial learning rate is set at 2e-5. We adopt a per-device training batch size of 12 and set the gradient accumulation steps to 4 for LLaMA-3.1-8B and Qwen3-8B. Meanwhile, for Phi-4 and Qwen2.5-7B, we use a per-device training batch size of 8 and set the gradient accumulation steps to 6.

\section{Prompt Templates}
\label{sec:prompt-template}
We present the two core prompt templates for our defense pipeline. Table~\ref{tab:cot_prompt} shows the CoT prompt used to identify critical evidence segments, while Table~\ref{tab:rewrite_prompt} guides the counterfactual rewriting process to generate contradictory alternatives.

\begin{table}[ht]
\begin{tcolorbox}[width=\columnwidth]
{
Answer the question based on the given passages. The following are the passages:\\
\textcolor[rgb]{0,0,0.9}{\{Context\}}\\
Answer the question based on the given passages.\\
Question: \textcolor[rgb]{0,0,0.9}{\{Question\}}\\
Please first provide your answer in the format of Answer:[Your answer]. Then provide your reasoning process step-by-step.(Only include explicit clues) \\
At the end of each reasoning step, include a new line that specifies the key information or reference content used in that step. \\
Please ensure that the [reference content] you include is the complete original sentence or consecutive sentences from the text. Please do not change the punctuation.  Do not use ellipses inside the sentence. \\
Follow this format:\\
Answer: [Your answer]\\
Step-by-step Reasoning:\\
1. [Reasoning step 1]\\{}[replaced by your reference content]\\
2. [Reasoning step 2]\\{}[replaced by your reference content]\\
}
\end{tcolorbox}
\caption{Chain-of-Thought Reasoning Prompt}
\label{tab:cot_prompt}
\end{table}

\begin{table}[t]
\begin{tcolorbox}[width=\columnwidth]
{

You are a creative contrarian. Given the question below, and the original answer, first propose a concise alternative answer—that is, a plausible but intentionally misleading answer. \\
Followed are some sentences supporting the original answer, please rewrite them. When rewriting each sentence, modify only the parts necessary to support the antifact answer. Parts unrelated to the answer must keep their original meaning. Be sure that the modified evidence sentences are sufficient to answer the original question. Output must be strictly in the specified JSON format, with no additional text.\\
\{\\
 "answer": "<your antifact answer here, just provide the answer phrase, no need for complete sentence>",\\
 "revised": [\\
 "<rewritten sentence 1>",\\
"<rewritten sentence 2>",\\
]\\
\}\\
Question:\\
\textcolor[rgb]{0,0,0.9}{\{Question\}}\\
Original answer:\\
\textcolor[rgb]{0,0,0.9}{\{Original Answer\}}\\
Sentences to rewrite:\\
\textcolor[rgb]{0,0,0.9}{\{Numbered Sentences\}}\\
}
\end{tcolorbox}
\caption{Counterfactual Evidence Rewriting Prompt}
\label{tab:rewrite_prompt}
\end{table}

\end{document}